\documentclass{article}

\usepackage[preprint]{neurips_2026}

\usepackage[utf8]{inputenc}
\usepackage[T1]{fontenc}
\usepackage{hyperref}
\usepackage{url}
\usepackage{booktabs}
\usepackage{amsfonts}
\usepackage{nicefrac}
\usepackage{microtype}
\usepackage{xcolor}

\usepackage{float}
\usepackage{amsmath}
\usepackage{graphicx}
\usepackage{booktabs,longtable,makecell,pdflscape}
\usepackage{tabularx}
\usepackage{array}
\usepackage{caption}

\title{From AutoResearch to Continual AutoResearch: Do Agents Learn Across Tasks?}

\author{%
  Jeremy Qin\textsuperscript{1,2}\thanks{Work done during an internship at RBC Borealis}
  \quad Ruizhi Deng\textsuperscript{1}
  \quad Mohamed Osama Ahmed\textsuperscript{1} \\
  \textbf{Frederick Tung\textsuperscript{1}
  \quad Leo Feng\textsuperscript{1}} \\[0.5em]
  \textsuperscript{1}RBC Borealis \\
  \textsuperscript{2}ELLIS Institute T\"ubingen, \\
  Max Planck Institute for Intelligent Systems, T\"ubingen AI Center
}

\begin{document}

\maketitle

\begin{abstract}
Autoresearch has shown that agents can autonomously write code, run experiments, and iteratively improve machine learning solutions. Yet most evaluations start a new agent session from scratch for each new problem, making them costly and leaving unclear whether agents can retain and apply prior insights when adapting to new tasks. We introduce \textsc{Continual AutoResearch}, an open-ended environment that compares persistent and reset instances of the same agent under matched task streams, data, and time budget. Continual AutoResearch covers nine task families, spanning synthetic tasks, symbolic regression, time-series forecasting, and tabular AutoML. Across five frontier agents, we measure first-solution quality, research efficiency, and final held-out performance. To distinguish insight transfer from session continuity, we run a memory ablation that vary what information agents retain from prior tasks. We show that persistent state improves research efficiency of all five agents, allowing them to find stronger solutions sooner. However, these gains do not consistently translate into better final solutions. Overall, our findings suggest that retaining experience and utilizing it effectively are distinct capabilities.

\end{abstract}

\section{Introduction} \label{sec:introduction}
Recently, there has been a growing interest in scaling frontier language models to different AI R\&D processes \citep{3692070.3692884, lu2026endtoend}. From optimizing a set of hyperparameters to post-training small language models, agents are becoming more autonomous, and more capable of tackling complex and lengthy tasks \citep{karpathy2026autoresearch, posttrainbench_2026}. As these agents shift towards longer horizon tasks, they increasingly accumulate insights, strategies, and practical experience. In real-world research, scientists often carry lessons from one project to the next, whether by reusing code or reapplying a specific post-training algorithm for example. We ask the question of whether experience acquired from one problem can benefit subsequent ones. Answering this requires understanding whether agents can retain knowledge across tasks and use it to adapt more effectively to new ones, bridging continual learning and meta-learning \citep{zhao2024expel,asawa2026continuallearningbenchevaluating,shu2026agentclrigorousevaluationcontinual}. We study this setting without updating model weights, in which case experience is transferred only through the state accumulated during execution.

\begin{figure*}[t]
    \centering
    \includegraphics[width=\textwidth]{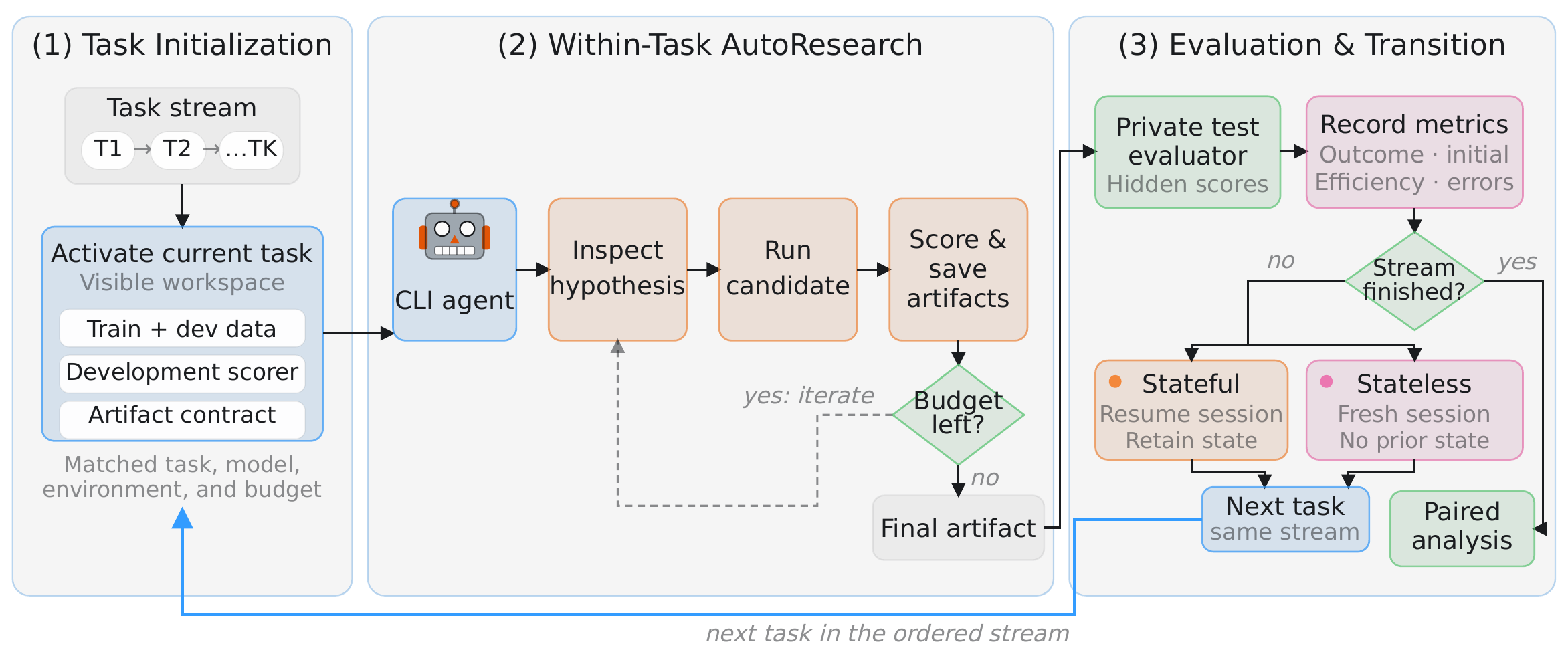}
    \caption{\textbf{Overview of  \textsc{Continual AutoResearch}.} Stateful and stateless agents process the same ordered task stream under matched environments, and per-task budgets. Each task consists of an autoresearch loop, but only the stateful agent retains experience across tasks. Private evaluation of final artifacts enables paired measurement of cross-task gains in performance, and efficiency.}
    \label{fig:framework}
\end{figure*}

Existing evaluations treat each R\&D problem in isolation by resetting the agent from scratch before each new task. Although autoresearch-style loops allow agents to use feedback to optimize solutions iteratively, the acquired experience is discarded once the task is finished. Consequently, these evaluations cannot assess if lessons about experimental design, modeling choices, or execution failures can be applied to related tasks and improve future research. While resetting prevents irrelevant or misleading experience from carrying over, it may also force agents to rediscover useful practices or repeat avoidable mistakes. However, retaining experience is not necessarily beneficial \citep{shu2026agentclrigorousevaluationcontinual}. Experience becomes useful only to the extent the tasks share transferable structure, such as common data-generating factors, modeling principles, or experimental approaches. Thus, whether autonomous research agents can identify and adapt such transferable experience to new tasks remains largely unevaluated. Figure~\ref{fig:framework} summarizes the paired evaluation pipeline.

We introduce \textsc{Continual AutoResearch}, an agent-agnostic evaluation environment for studying whether agents can retain and use prior ML experience while adapting to new research tasks. We compare persistent and reset instances of the same fixed model under identical tasks, data, and time budgets. Our environment combines controlled synthetic tasks, symbolic regression, time series forecasting and tabular ML problems. It measures initialization, search efficiency, and outcome gains. However, a stateful session incorporates multiple potential sources of improvement, such as transcript history, code reuse, procedural familiarity, and scientific insight. \textsc{Continual AutoResearch} contributes, to our knowledge, the first study of a continual AI R\&D setting analyzing whether agents can efficiently adapt to new research tasks.

\textbf{Contributions} Our main contributions are:
\begin{itemize}
    \item We introduce \textsc{Continual AutoResearch}, an open-ended environment for evaluating whether agents continually adapt by applying prior experience to new ML research tasks. We pair persistent and reset runs on matched task streams spanning nine task families across synthetic tasks, symbolic regression, time series forecasting, and tabular AutoML.
    \item We evaluate five frontier CLI agents on held-out performance, initial-solution quality and research efficiency. Persistent state improves mean fixed-budget research efficiency for all five agents by helping them produce strong solutions sooner, but these gains do not consistently translate into better final outcomes.
    \item We find that task-relevant insights change initialization and search behavior, but do not consistently improve final performance through interventions with matched agent-authored insights, shuffled insights, and no memory. Retaining relevant knowledge and using it effectively are therefore separate capabilities.
\end{itemize}

\section{Related Work} \label{sec:rel_works}

\subsection{Automated AI R\&D}

Automated AI R\&D studies whether agents can do research with limited supervision. Karpathy's autoresearch \citep{karpathy2026autoresearch} introduces a minimal loop in which an agent edits a training script, evaluates the result, and keeps or discards the change. PostTrainBench \citep{posttrainbench_2026}, InferenceBench \citep{yeon2026inferencebenchbenchmarkopenendedllm}, and Claudini \citep{panfilov2026claudiniautoresearchdiscoversstateoftheart} extend local optimization to LLM post-training, inference systems, and adversarial attack discovery. AutoLab \citep{xu2026autolabfrontiermodelssolve} broadens evaluation to long-horizon closed-loop optimization across research and engineering tasks, while AutoScientists \citep{gao2026autoscientistsselforganizingagentteams} studies self-organizing agent teams that share experimental state during long-running scientific search. AutoResearchBench \citep{xiong2026autoresearchbenchbenchmarkingaiagents} focuses on multi-step scientific literature discovery, and AutoResearch AI \citep{tie2026autoresearchaiaipoweredresearch} surveys automation across the broader research workflow. However, these works primarily measure progress within a single research problem. We instead ask whether experience transfers across ML tasks and measure its effect on initialization, search efficiency, and final selection.

\subsection{Continual Learning in Language Agents}
Continual-learning benchmarks evaluate whether language agents can use experience from earlier tasks. Earlier systems such as ExpeL \citep{zhao2024expel} study extracting and reusing textual experience across tasks, while LifelongAgentBench \citep{zheng2025lifelongagentbenchevaluatingllmagents} evaluates knowledge accumulation over interdependent tasks in interactive environments. Continual Learning Bench \citep{asawa2026continuallearningbenchevaluating} spans software engineering, signal processing, epidemiological forecasting, database querying, strategic games, and demand forecasting. AgentCL \citep{shu2026agentclrigorousevaluationcontinual} constructs controlled streams across coding, deep research, and language reasoning to evaluate memory and cross-task composition. FutureSim \citep{goel2026futuresimreplayingworldevents} studies world-event forecasting from chronologically revealed news, while SWE-Bench-CL \citep{joshi2025swebenchclcontinuallearningcoding} organizes software-engineering issues into repository-level sequences. Continual AutoResearch differs in what is being evaluated: every task is an iterative, executable ML research loop in which an agent forms modeling hypotheses, edits and runs code, generates multiple solutions, and selects a final solution under a fixed budget. We measure how prior state changes initialization, research exploration, and final outcome across a sequence of these loops.

\section{Framework} \label{sec:framework}
\subsection{Setup}
We formulate continual AI R\&D as an ordered sequence of research tasks $\mathcal{T}=(T_1,\ldots,T_K)$. A task $T_i$ consists of a task description and instructions, training and development data, and a fixed wall-clock budget $B_i$. Within $B_i$, the agent can experiment without restrictions. The agent can submit multiple solutions and need to select a final one before the budget expires. All submissions are then evaluated on a hidden test set after the run. To prevent the agent from reward hacking, intermediate submissions test scores are hidden. The environment does not provide a specific harness for planning or memory management. Agents operate through their native CLI interfaces and can modify files, execute code, and inspect visible feedback autonomously. Models and budget values are specified in the experimental setup described in Section~\ref{sec:results}.

\subsection{Stateful and Stateless Agents}
We compare two evaluation settings: stateful execution, in which the same agent instance persists across tasks, and stateless execution, in which each task resets with a fresh instance of the same agent. Let $s_i$ denote the agent's working state when it begins task $T_i$, including its session history, files,
experimental results, and execution experience. Under stateful execution, $s_i$ may include information retained from tasks $T_1,\ldots,T_{i-1}$; under stateless execution, the working state is reset before each task and contains no information from earlier tasks.

\textbf{Stateful execution.} We resume the same native agent session and workspace across the stream. Conversation history, files, experimental results, execution traces can then carry from one task to the other. Feedback from earlier tasks remains visible in this state. We do not provide an additional harness-written memory or summary. Everything is handled through their native CLI harness.

\textbf{Stateless execution.} We launch a new agent session and clean workspace for every task. No transcript, files, code, or execution state from earlier tasks is available. We otherwise match the task, model, environment, wall-clock budget, and evaluation procedure to the stateful run. Since task $T_1$ doesn't have access to any prior sessions, our transfer results are computed over $T_2,\ldots,T_K$.

\subsection{Measuring Cross-Task Improvement}
We use different metrics for different tasks following their original setup. This includes accuracy, F1 for classification, RMSE for time series, and $R^2$ for symbolic regression. Following the normalized-score approach used by SELA \citep{chi2024selatreesearchenhancedllm}, we normalize each metric between $[0,1]$, where higher is better. This transformation allows aggregate gains to be comparable across task families and allows the framework to include tasks with different evaluation metrics. Let $m$ denote a raw metric value and $\mu$ its metric type. The transformation is:

\begin{equation}
r(m;\mu) =
\begin{cases}
m, & \mu \in \{\mathrm{Accuracy},\mathrm{F1}\}, \\
(1+m)^{-1}, & \mu = \mathrm{RMSE}, \\
\bigl(2-\min\{m,1\}\bigr)^{-1}, & \mu = R^2.
\end{cases}
\end{equation}

Accuracy and F1 are already bounded between 0 and 1, so they are left unchanged. RMSE is a lower-is-better metric that can take on any positive real value; $(1+m)^{-1}$ maps zero error to one and larger errors toward zero. RMSE is standardized using the standard deviation of each channel's training set before the transformation. $R^2$ is higher-is-better, bounded above by one, and unbounded below; $(2-\min\{m,1\})^{-1}$ maps one to one and increasingly negative values toward zero. These transformations preserve the ordering of solutions within each task while placing every metric on a common scale. We do not average unnormalized raw metrics across heterogeneous tasks. For any higher-is-better measure $q$, let $M^q_{i,c}$ denote its value on task $T_i$ under setting $c$. We define its paired gain in normalized reward points as

\begin{equation}
\Delta_i^q = 100\left(
M^q_{i,\mathrm{stateful}}-
M^q_{i,\mathrm{stateless}}
\right).
\end{equation}

A positive gain indicates that stateful execution performs better. We report gains for three complementary metrics that capture different parts of the research process: final selected reward ($q=\mathrm{out}$), first-solution reward ($q=\mathrm{init}$), and fixed-horizon search utility ($q=\mathrm{search}$). These metrics respectively measure the quality of the final solution, the quality of the agent's initial solution, and its ability to find and explore strong solutions efficiently within the task budget. For aggregate results, we first average gains within each task stream and then average across streams.

Final performance alone does not reveal how prior experience changes the research process. In practice, agents run under limited time and compute budgets, making both the quality of their first solution and the speed at which they explore other strong solutions important. We define initialization quality as the hidden-test reward of the first valid submission and use the corresponding initialization gain $\Delta_i^{\mathrm{init}}$ to compare the two settings. To measure efficiency over the full budget, the evaluator scores every valid submission after the run and tracks the best reward achieved by each point in time. For setting $c$, let $r_{ijc}$ be the hidden reward of attempt $j$ completed at time $t_{ijc}$, and let $B_i$ be the task budget. The normalized area under this best-so-far curve defines the fixed-horizon search utility:

\begin{equation}
U_{i,c} = \frac{1}{B_i}\int_0^{B_i}
\max_{j:t_{ijc}\leq t} r_{ijc}\,dt.
\end{equation}

Before the first valid submission, we set the curve to zero. If a run ends early, its last value is carried to the end of the budget. Higher utility therefore means that better solutions are found earlier: the metric jointly captures solution quality and time of discovery. We set $M^{\mathrm{search}}_{i,c}=U_{i,c}$ and call $\Delta_i^{\mathrm{search}}$ the time-convergence gain. Our primary transfer metric is its mean over $i=2,\ldots,K$; initialization and outcome gains provide complementary views of the first valid and final selected solutions. The curve is constructed only after the run and is never shown to the agent in case it tries to reward hack.

\section{Settings} \label{sec:settings}
\subsection{Design Rationale}

We evaluate agents on ordered task streams whose tasks share reusable structure but differ in their target, dataset, or sampling process. This prevents the solution of the previous problem to be resubmitted directly by the agent. Instead, agents must identify what transfers, re-estimate what is task-specific, and use model feedback to verify earlier assumptions. The settings progress from controlled synthetic tasks with known recurring structure to realistic tasks with similar methodological approaches. This allows us to study transfer progressively, from recurring scientific structure to more general research practices. We report public dataset names for clarity, but agents receive neutral identifiers during evaluation to prevent the agent of recognizing the dataset and taking shortcuts. Table~\ref{tab:continual-automl-settings} summarizes the four evaluation settings and nine task families. Appendix~\ref{app:setting-details} describes the construction of each stream in detail and provides information needed to reproduce them.

\subsection{Evaluation Settings}

\paragraph{Synthetic AutoML.}
Synthetic AutoML models a setting where the same underlying dataset is reused across multiple tasks, as often occurs in organizations that build different models from a shared pool of data. The setting is composed of three streams that contain five classification tasks with different target rules. In the shared-latent stream, observed features are noisy mixtures of recurring latent factors, so representation discovery can transfer while the predictive factors change. In the shared-sparse stream, each target depends on a different subset of a small pool of relevant features, making feature-selection and regularization strategies transferable even though the exact predictive features change across tasks. In the shared-regime stream, a recurring regime variable affects each target differently, rewarding agents that retest regime-dependent interactions rather than purely relying on previous solutions. 

\begin{table}[t]
\centering
\small
\renewcommand{\arraystretch}{1.35}

\begin{tabularx}{\linewidth}{
  @{}
  >{\raggedright\arraybackslash}p{0.30\linewidth}
  !{\vrule width 0.5pt}
  >{\raggedright\arraybackslash}X
  @{}
}
\toprule
\textbf{Setting} & \textbf{Task families} \\
\midrule

Synthetic AutoML (3)
& Shared latent, shared sparse, shared regime \\
\hline

Symbolic regression (1)
& Canonical coordinates \\
\hline

Time-series forecasting (3)
& Energy systems, traffic networks, ETT cross-resolution \citep{10.1145/3209978.3210006,godahewa2021monash,DBLP:conf/aaai/SongLGW20,zhou2021aaai-informer} \\
\hline

Tabular AutoML (2)
& SELA evaluation subset (AMLB/Kaggle), MLE-Bench Lite \citep{chi2024selatreesearchenhancedllm,JMLR:v25:22-0493,chan2025mlebench} \\

\bottomrule
\end{tabularx}

\caption{
\textbf{\textsc{Continual AutoResearch} evaluation coverage.}
The evaluation spans four settings and nine task families, ranging from controlled synthetic and symbolic tasks with known recurring structure to real-world forecasting and tabular AutoML streams.
}
\label{tab:continual-automl-settings}
\end{table}

\paragraph{Symbolic regression.}
Symbolic regression is an interesting setting that tests transfer of a scientific search strategy rather than a fitted predictor. The five-task canonical-coordinates stream contains equations that become simpler after an appropriate change of variables, including centering, rescaling, or combining variables into a radial coordinate. We also include an asymmetric control task to test whether the agent verifies a previously useful symmetry instead of applying it automatically. Given noisy training and development samples, the agent submits an explicit expression that is evaluated on private interpolation and extrapolation samples using mean \(R^2\).

\paragraph{Time-series forecasting.}
Time-series forecasting provides three levels of naturally occurring similarity. The ETT cross-resolution stream \citep{zhou2021aaai-informer} varies station and sampling frequency within the same electricity-transformer system, requiring periods and context lengths to be recalibrated. The PEMS stream \citep{DBLP:conf/aaai/SongLGW20} moves across traffic sensor networks and time periods, preserving domain-level seasonality while changing channel relationships. The energy-system stream \citep{10.1145/3209978.3210006, godahewa2021monash} spans broader sources and sampling rates, allowing for more open-ended transfer across tasks. Across the streams, agents can reuse procedures for normalization, lag construction, periodicity detection, seasonal baselines, and chronological validation, but must avoid overrelying on prior solutions. For evaluation, agents submit executable forecasts, which are then run on hidden test splits using RMSE.

\paragraph{Tabular AutoML.}
Tabular AutoML uses a five-task subset of the datasets used in SELA's evaluation \citep{chi2024selatreesearchenhancedllm}, drawn from AMLB \citep{JMLR:v25:22-0493} and Kaggle competitions, and a three-task slice of MLE-Bench Lite \citep{chan2025mlebench}. We use fixed subsets so that the full task stream remains feasible to run across all agents while preserving the same task order and per-task budget. Unlike the controlled settings, each task uses a different real-world dataset with its own characteristics and requirements. These streams test whether agents can transfer useful research practices across datasets while adapting to the specific requirements of each new task.

\begin{figure}[t]
    \centering
    \includegraphics[width=\textwidth]{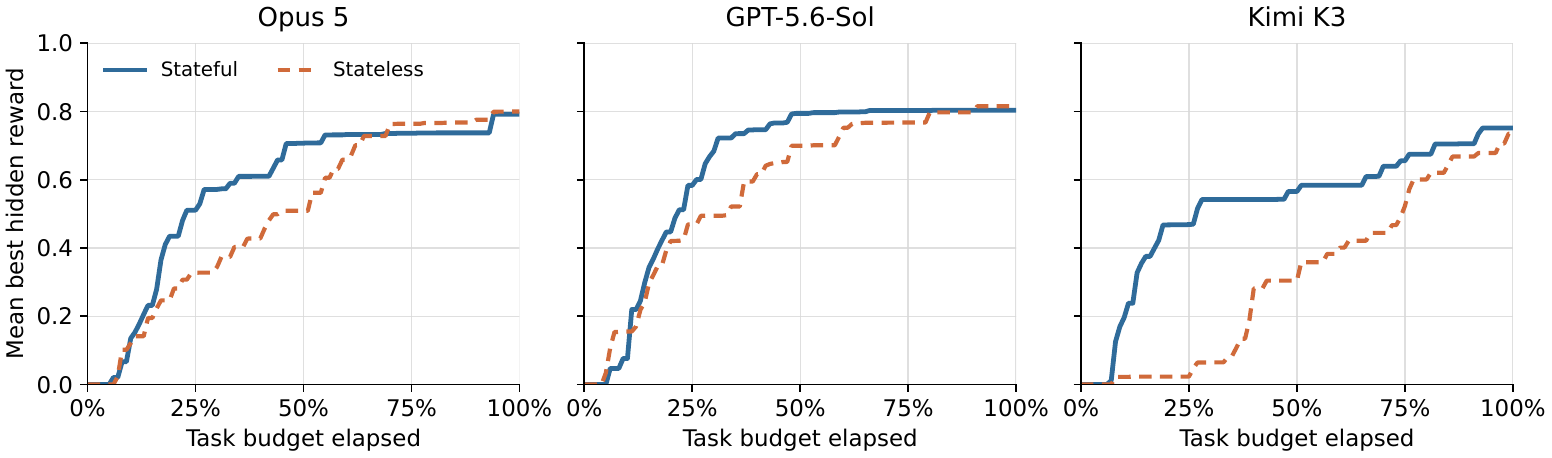}
    \caption{\textbf{Persistent-state search advantage over time.} Mean best hidden reward as a function of normalized task-budget elapsed. Task-level best-so-far curves are constructed post hoc, averaged within each family, and then averaged with equal family weight. Persistent execution generally finds useful solutions earlier, although the size and duration of the advantage vary across agents.}
    \label{fig:search-trajectories}
\end{figure}

\section{Results} \label{sec:results}
\subsection{Experimental Setup}
We evaluate five frontier agents via their native CLI harnesses: Sonnet 4.6 and Opus 5 through Claude Code, GPT-5.4 and GPT-5.6-Sol through Codex, and Kimi K3 through OpenCode. The evaluation covers four task domains and nine task families, each of which contains three to five tasks. The task budget $B_i$ is set to default to 30 minutes unless otherwise specified. For each ordered stream, we compare a stateful run with reset runs using the same exact setup. 
To assess run-to-run variability, we repeated the matched stateful--reset comparison three times for GPT-5.6-Sol (Appendix Figure~\ref{fig:stochasticity}). Outcome and initialization gains varied by at most 0.75 points, and time-convergence gains by at most 7.14, both small relative to the aggregate effects. Since repeating the full benchmark is costly, all other comparisons use a single matched run pair.

\subsection{Persistent State Improves Research Efficiency}
Mean time-convergence gain is positive for all five agents: \(+21.3\) normalized reward points for Kimi K3, \(+20.9\) for Sonnet 4.6, \(+7.5\) for Opus 5, \(+6.1\) for GPT-5.6-Sol, and \(+3.0\) for GPT-5.4 (Table~\ref{tab:model-results-aggregate}). Time-convergence measures fixed-budget research efficiency using the highest hidden-test reward generated by each submission within the budget, with hidden scores computed post hoc. This metric measures both speed and quality: a solution contributes more the higher its reward and the earlier it is produced. The reward trajectories (Figure~\ref{fig:search-trajectories} show that persistent agents produce higher reward solutions sooner, while reset agents often close the gap near the deadline. To determine how long this advantage persists, we compare the best reward achieved at 25\%, 50\%, 75\%, and 100\% of the budget (Appendix Table~\ref{tab:budget-checkpoints}). Stateful execution leads for all five agents at 25\% and 50\%, but the mean gap decreases from \(+24.0\) points at 25\% to \(+1.2\) at the deadline, where only two agents remain positive. Persistence therefore only shows consistent advantage early in the research process. On the controlled synthetic streams, this advantage increases with task position (Figure~\ref{fig:synthetic-task-position-summary}). For all models except Sonnet 4.6, the mean advantage across the synthetic task families is larger on every later task than on Task 1, suggesting that accumulated experience becomes more useful as the stream progresses.

To determine what drives the time-convergence gains, Table~\ref{tab:time-convergence-robustness} reports time to the first valid solution, first-solution quality, valid candidate counts, and gains after controlling for candidate count and first-solution latency. Stateful execution reaches the first valid solution sooner for every agent, with the largest reductions for Kimi K3 (\(9.2\) minutes) and Sonnet 4.6 (\(8.6\) minutes). It also produces fewer valid candidates for four of five agents. The gain is then not explained by producing more candidates; invalid, unchanged, or lower-reward submissions do not improve the best-so-far curve. First-solution quality is also higher for four agents, most clearly Sonnet 4.6 (\(+6.8\) points) and GPT-5.4 (\(+3.7\)). Sonnet benefits from both faster initialization and a stronger first solution, whereas Kimi’s large full-budget gain is driven mainly by reaching its first valid solution sooner.

Matching candidate counts reduces the gain for every agent. Sonnet keeps a \(+6.5\)-point advantage, Kimi K3 and GPT-5.4 retain smaller gains, and Opus 5 and GPT-5.6-Sol fall near zero. Starting each trajectory at its first valid solution produces the same pattern, although this control still preserves differences in first-solution quality. Under a stricter control that also subtracts each run's first-valid reward, the gain becomes negative for every agent (Appendix~\ref{app:time-convergence-robustness}). Persistent state therefore improves research efficiency by producing a valid solution sooner or by producing a stronger first solution. The current results do not show a general improvement in optimization after initialization.

\begin{table}[t]
\centering
\caption{Model-level results over 32 transfer tasks per agent. Reset outcome is the family-balanced mean final reward under reset execution. Gains are stateful-minus-reset normalized reward points reported as mean [median]. Win rates report the family-balanced fraction of tasks on which stateful execution outperforms reset execution.}
\label{tab:model-results-aggregate}
\begin{small}
\setlength{\tabcolsep}{2.5pt}
\renewcommand{\arraystretch}{1.08}
\begin{tabular*}{0.98\textwidth}{@{\extracolsep{\fill}}lcccccc@{}}
\toprule
Model & \makecell{Stateless\\outcome} & \makecell{Outcome\\gain} & \makecell{Init.\\gain} & \makecell{Time-conv.\\gain} & \makecell{Outcome\\win rate} & \makecell{Init.\\win rate} \\
\midrule
Opus 5      & 79.9 & -0.9 [+0.0] & +1.3 [+0.5] & +7.5 [+10.3]  & 34.3\% & 46.3\% \\
GPT-5.6-Sol & 80.5 & -0.8 [+0.2] & -1.0 [+1.2] & +6.1 [+7.4]  & 59.3\% & 69.4\% \\
Kimi K3     & 73.1 & +0.4 [+2.0] & +1.8 [+2.0] & +21.3 [+27.5] & 62.0\% & 64.8\% \\
Sonnet 4.6  & 70.3 & +7.6 [+1.3] & +7.0 [+7.0] & +20.9 [+20.6] & 66.7\% & 75.0\% \\
GPT-5.4     & 72.9 & +0.2 [-0.2] & +3.0 [+3.1] & +3.0 [+2.0]   & 51.9\% & 75.0\% \\
\bottomrule
\end{tabular*}
\end{small}
\end{table}

\subsection{Faster Search Does Not Consistently Improve Final Performance}
Final outcome gains are less consistent than time-convergence gains. Sonnet 4.6 is the only agent with a substantial mean final outcome gain, at \(+7.6\) points. Kimi K3 and GPT-5.4 remain near zero at \(+0.4\) and \(+0.2\) points, while GPT-5.6-Sol and Opus 5 decline by \(-0.8\) and \(-0.9\) points. We also report outcome win rate which is the proportion of transfer tasks on which the stateful final solution outperforms the reset solution. Unlike outcome gain, it measures the consistency of transfer across tasks rather than its average magnitude. It ranges from \(34.3\%\) for Opus to \(66.7\%\) for Sonnet.

Faster search also does not necessarily improve the final solution. A reset run may reach a solution of comparable quality later in the task budget, or an agent may generate a strong candidate and select a weaker one that overfits. We measure the latter with selection regret: the best hidden reward generated during a run minus the hidden reward of its selected solution. A positive stateful-minus-reset difference means that stateful execution has greater selection regret. This difference ranges from \(-0.7\) to \(+1.3\) points across agents and remains within \(\pm0.8\) points for four of five agents. Selection failures can be severe on individual tasks (Appendix Figure~\ref{fig:failure-forecast-selection}), but they do not explain the aggregate difference between search efficiency and final outcomes.

Transfer gains also depend on model capability. Opus 5 and GPT-5.6-Sol have the strongest reset performance, but their initialization and outcome gains are smaller than those of Sonnet 4.6 and GPT-5.4, respectively (Appendix Figure~\ref{fig:absolute-outcome-by-agent}). Sonnet 4.6 obtains \(+7.0\) initialization and \(+7.6\) outcome gains, compared with \(+1.3\) and \(-0.9\) for Opus 5. GPT-5.4 similarly obtains \(+3.0\) and \(+0.2\), compared with \(-1.0\) and \(-0.8\) for GPT-5.6-Sol. More capable models may already produce strong solutions from a reset state, leaving less room to benefit from persistence. However, model capability does not consistently predict time-convergence gain. Transfer also varies by task family, with no family except SELA AutoML improving uniformly across models (Appendix Figure~\ref{fig:outcome-gain-by-family}).

\subsection{Time Budget Changes the Value of Persistence}
\label{subsec:budget-ablation}
\begin{figure}[t]
    \centering
    \includegraphics[width=\textwidth]{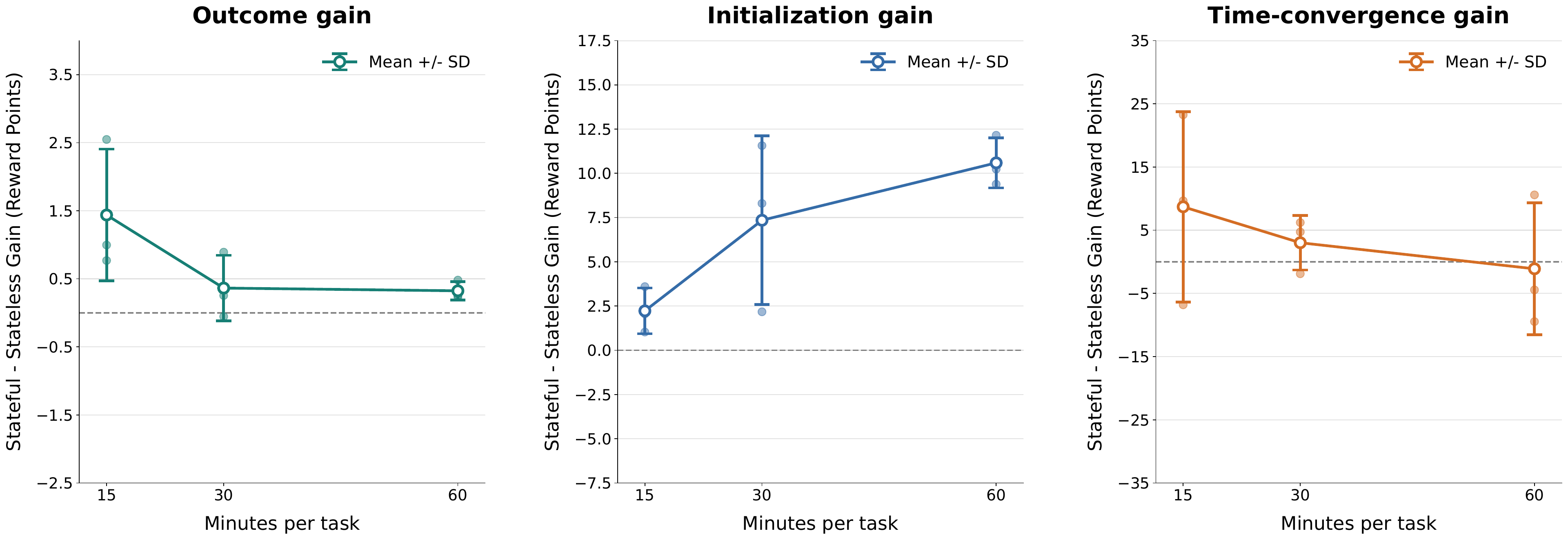}
    \caption{\textbf{Effect of per-task wall-clock budget on Shared Regime.} Stateful-minus-reset final-outcome, initialization, and time-convergence gains over transfer tasks (positions 2–5), in normalized reward points. Points show matched repetitions; markers and bars show their mean and sample standard deviation across three repetitions. Each budget is a separate run, rather than a checkpoint taken from a longer run.}
    \label{fig:time-ablation}
\end{figure}

Figure~\ref{fig:time-ablation} compares stateful and reset execution of GPT-5.6-Sol on Shared Regime under three different time budget: 15min, 30min and 60min. As the time budget increases, the mean final outcome gain becomes smaller while the mean initialization gain grows. We also observe a decrease in time-convergence gains, which is consistent with what we would expect of a reset run narrowing an early stateful advantage through within-task exploration. Conversely, longer preceding tasks may give the stateful agent more experience to transfer to its first solution. Together, these observations emphasize the tradeoff of the different gains with respect to the task time horizon. 

\subsection{Explicit Memory Changes Search but Is Not Reliably Beneficial}
To separate explicit knowledge transfer from uninterrupted session state, we run a memory ablation on the Energy-systems stream with three conditions: matched memory, shuffled memory, and no memory. Every task starts with a fresh agent session, and information can transfer only through insight cards. These cards are agent-authored summaries of earlier experiments that record claims, supporting evidence, confidence, scope, recommended actions, and failure conditions. In the matched-memory condition, before each task, the agent receives the full set of cards accumulated from earlier tasks in the same stream. In the shuffled-memory condition, it follows a similar setup, but the cards come from an unrelated stream. In the no-memory condition, it receives no cards. Comparing matched with shuffled memory tests whether the relevance of the transferred content matters, while comparison with no memory tests the effect of providing cards at all.

Relevant cards don't consistently result in gains. Compared with shuffled cards, matched cards improve initialization for Opus 5 (\(+2.6\) points), GPT-5.4 (\(+5.9\)), and GPT-5.6-Sol (\(+3.9\)), but reduce time-convergence for all three. Sonnet shows the reverse pattern where initialization declines by \(2.9\) points while time-convergence improves by \(25.6\). Final matched-versus-shuffled outcome effects are small for Sonnet, Opus, and GPT-5.4, and negative for GPT-5.6-Sol. Matched and shuffled cards also produce comparable final gains over no memory for Sonnet and GPT-5.4. These results show that relevant information about prior experiments affect initialization and search efficiency but do not consistently improve final performance. (Appendix~\ref{app:memory-contrasts}).

The logged agent traces also illustrate how persistent memory change agent behavior. Stateful Sonnet reuses and recalibrates a forecasting pipeline; matched cards lead Codex to adapt a seasonal-residual hypothesis to a new sampling rate; and highly specific cards anchor Sonnet to a weaker model family (Appendix~\ref{app:agent-traces}). The failure cases further show agents replacing robust candidates with overfitted ones or applying an earlier operational lesson too strongly (Appendix~\ref{app:agent-failures}). These cases are good examples of why reuse, verification, exploration, and final selection must be evaluated separately.

\section{Discussion} \label{sec:discussion} 
\subsection{What Persistence Changes}

Persistent state improves mean fixed-budget time-convergence for all five agents, primarily by helping them reach valid solutions earlier and, for some agents, with higher initial quality. Sonnet benefits from both, whereas several other agents benefit mainly from earlier initialization (Table~\ref{tab:time-convergence-robustness}). Only Sonnet retains a clear advantage after matching candidate counts and excluding the latency before the first valid solution. Once each run’s first-solution reward is also subtracted, no agent retains an advantage. These process gains do not consistently translate into better final performance. 

The benefit of persistence also varies across the stream. The task-position analysis shows larger search gains on later synthetic tasks, suggesting that useful experience can accumulate. However, variation across task families and the memory ablation show that additional history is not automatically beneficial. Within the Claude and GPT model families, agents with lower reset performance have more room for initialization and outcome gains, while time-convergence gains follow a different pattern. Persistence is therefore not uniformly beneficial across agents, tasks, or metrics.

\subsection{Implications for Continual Research Agents}
Overall, our findings favor persistent execution when compute or time budgets are limited. In the time-horizon ablation, the gains from persistence for both final outcome and time to convergence decrease as budgets grow. This is consistent with our observation of reset agents partially catching up through within-task optimization. Persistence is therefore most valuable when budgets are tight, where its advantage narrows as they grow.

Our results also identify two ways agents can fail to adapt effectively across tasks. First, agents can apply an earlier lesson too broadly, narrowing their search around assumptions that do not hold on the current task. Second, agents can find a strong candidate early but later replace it using unreliable development evidence (Appendix~\ref{app:agent-failures}). These observations do not suggest a single memory or search policy, but they show that retaining experience is insufficient without mechanisms for verifying and revising it. These risks may become more important for deployed agents operating across longer task sequences. An incorrect assumption, brittle code path, or obsolete instruction retained in a workspace can influence multiple later tasks. When state includes externally supplied artifacts, persistent reuse could also allow misleading or malicious content introduced during one task to affect future behavior. We do not evaluate adversarial manipulation in this work, but the observed failures motivate testing safeguards such as provenance tracking, explicit applicability conditions, and evidence-based memory revision. They also motivate process-level evaluation: final performance alone cannot distinguish late candidate discovery, execution failures, weak candidate generation, and incorrect final selection.

\subsection{Limitations and Future Work} 

We evaluate each model through its native CLI scaffold, so stateful and reset comparisons are matched within each agent, but cross-agent differences may reflect both the model and its scaffold. Our ablations also remain limited in scope. Future work should separately test insights, code, transcripts, and execution state, together with explicit mechanisms for selecting or forgetting memories. More broadly, our streams contain only three to five tasks with 30-minute budgets. It would be interesting to scale to streams with hundred of tasks, or with tasks that require much longer time budget such as LLM post-training like PostTrainBench \citep{posttrainbench_2026} who are of the order of 10 hours. These experiments would test further mechanisms that are less studied in our work such as compaction, etc. Another interesting direction along this line of work is to study continual learning in a multi-agent system. At larger scale, research is also likely to be distributed across multiple agents working in parallel similar to a research team or research company. Continual improvement is comparatively straightforward to measure for a single agent. In a multi-agent system, however, what each agent learns and how that knowledge spreads depend on the interactions between agents and the topology connecting them, which makes cross-task improvement harder to attribute and evaluate.

\subsection{Conclusion}

We introduced Continual AutoResearch, an environment for evaluating research agents across related tasks rather than in isolation.  Across five frontier agents and nine task families, persistent state improves mean fixed-budget research efficiency by helping agents produce stronger solutions sooner, with the balance between speed and quality varying by agent. Sonnet 4.6 is the only agent that retains a substantial advantage under the candidate-matched and post-first timing controls and obtains a substantial aggregate final-outcome gain. Retaining experience and leveraging it well are therefore separate capabilities: prior code, hypotheses, and procedures can accelerate and strengthen initialization, while weak follow-up search, anchoring, or poor final selection can erase the benefit. Effective continual research requires selective memory, explicit tests of transferred assumptions, continued exploration of alternatives, and robust candidate selection.

\section*{Ethics Statement}
We propose a new environment for studying AI agents that carry experience across research tasks, a capability we show could accelerate automated machine learning research. The long-horizon autonomy of the agents naturally introduces certain risks such as
unmonitored code execution, compute usage,  credential leakage, etc. We log the agents' edits, executions, and submissions for post-hoc analysis. Empirically, we report performance and failures so that failure modes of agent trajectories remain visible. We encourage active monitoring of autonomous agents' reasoning and actions to ensure safe model behavior.

\section*{AI Use Statement}
We used LLMs to assist with code development, figure plotting, table formatting and writing polishing for grammar and syntax errors. All generated outputs were reviewed and revised by the authors.

\section*{Reproducibility Statement}
All datasets, prompts, instructions, and others are provided in the paper and Appendix for more details. The code will be released alongside the camera-ready version to reproduce all results mentioned in the paper. We note that agents are inherently stochastic.
\bibliographystyle{plainnat}
\bibliography{references}

@inproceedings{posttrainbench_2026,
  title     = {PostTrainBench: Can LLM Agents Automate LLM Post-Training?},
  author    = {Ben Rank and Hardik Bhatnagar and Ameya Prabhu and Shira Eisenberg and Karina Nguyen and Matthias Bethge and Maksym Andriushchenko},
  booktitle = {International Conference on Machine Learning (ICML)},
  year      = {2026},
  eprint    = {2603.08640},
  archivePrefix = {arXiv},
  primaryClass  = {cs.SE},
  url       = {https://arxiv.org/abs/2603.08640}
}

@misc{yeon2026inferencebenchbenchmarkopenendedllm,
      title={InferenceBench: A Benchmark for Open-Ended LLM Inference Optimization by AI Agents}, 
      author={Jehyeok Yeon and Ben Rank and Maksym Andriushchenko},
      year={2026},
      eprint={2607.20468},
      archivePrefix={arXiv},
      primaryClass={cs.AI},
      url={https://arxiv.org/abs/2607.20468}, 
}

@misc{panfilov2026claudiniautoresearchdiscoversstateoftheart,
      title={Claudini: Autoresearch Discovers State-of-the-Art Adversarial Attack Algorithms for LLMs}, 
      author={Alexander Panfilov and Peter Romov and Igor Shilov and Yves-Alexandre de Montjoye and Jonas Geiping and Maksym Andriushchenko},
      year={2026},
      eprint={2603.24511},
      archivePrefix={arXiv},
      primaryClass={cs.LG},
      url={https://arxiv.org/abs/2603.24511}, 
}

@misc{asawa2026continuallearningbenchevaluating,
      title={Continual Learning Bench: Evaluating Frontier AI Systems in Real-World Stateful Environments}, 
      author={Parth Asawa and Christopher M. Glaze and Gabriel Orlanski and Ramya Ramakrishnan and Benji Xu and Asim Biswal and Vincent Sunn Chen and Frederic Sala and Matei Zaharia and Joseph E. Gonzalez},
      year={2026},
      eprint={2606.05661},
      archivePrefix={arXiv},
      primaryClass={cs.AI},
      url={https://arxiv.org/abs/2606.05661}, 
}

@misc{shu2026agentclrigorousevaluationcontinual,
      title={AgentCL: Toward Rigorous Evaluation of Continual Learning in Language Agents}, 
      author={Yiheng Shu and Bernal Jiménez Gutiérrez and Saisri Padmaja Jonnalagedda and Yuguang Yao and Huan Sun and Yu Su},
      year={2026},
      eprint={2606.02461},
      archivePrefix={arXiv},
      primaryClass={cs.AI},
      url={https://arxiv.org/abs/2606.02461}, 
}

@misc{goel2026futuresimreplayingworldevents,
      title={FutureSim: Replaying World Events to Evaluate Adaptive Agents}, 
      author={Shashwat Goel and Nikhil Chandak and Arvindh Arun and Ameya Prabhu and Steffen Staab and Moritz Hardt and Maksym Andriushchenko and Jonas Geiping},
      year={2026},
      eprint={2605.15188},
      archivePrefix={arXiv},
      primaryClass={cs.LG},
      url={https://arxiv.org/abs/2605.15188}, 
}

@misc{joshi2025swebenchclcontinuallearningcoding,
      title={SWE-Bench-CL: Continual Learning for Coding Agents}, 
      author={Thomas Joshi and Shayan Chowdhury and Fatih Uysal},
      year={2025},
      eprint={2507.00014},
      archivePrefix={arXiv},
      primaryClass={cs.LG},
      url={https://arxiv.org/abs/2507.00014}, 
}

@misc{chi2024selatreesearchenhancedllm,
      title={SELA: Tree-Search Enhanced LLM Agents for Automated Machine Learning}, 
      author={Yizhou Chi and Yizhang Lin and Sirui Hong and Duyi Pan and Yaying Fei and Guanghao Mei and Bangbang Liu and Tianqi Pang and Jacky Kwok and Ceyao Zhang and Bang Liu and Chenglin Wu},
      year={2024},
      eprint={2410.17238},
      archivePrefix={arXiv},
      primaryClass={cs.AI},
      url={https://arxiv.org/abs/2410.17238}, 
}

@inproceedings{
chan2025mlebench,
title={{MLE}-bench: Evaluating Machine Learning Agents on Machine Learning Engineering},
author={Jun Shern Chan and Neil Chowdhury and Oliver Jaffe and James Aung and Dane Sherburn and Evan Mays and Giulio Starace and Kevin Liu and Leon Maksin and Tejal Patwardhan and Aleksander Madry and Lilian Weng},
booktitle={The Thirteenth International Conference on Learning Representations},
year={2025},
url={https://openreview.net/forum?id=6s5uXNWGIh}
}

@inproceedings{zhou2021aaai-informer,
  title     = {{Informer: Beyond Efficient Transformer for Long Sequence Time-Series Forecasting}},
  author    = {Zhou, Haoyi and Zhang, Shanghang and Peng, Jieqi and Zhang, Shuai and Li, Jianxin and Xiong, Hui and Zhang, Wancai},
  booktitle = {AAAI Conference on Artificial Intelligence},
  year      = {2021},
  pages     = {11106-11115},
  doi       = {10.1609/AAAI.V35I12.17325},
  url       = {https://mlanthology.org/aaai/2021/zhou2021aaai-informer/}
}

@inproceedings{DBLP:conf/aaai/SongLGW20,
  author       = {Chao Song and
                  Youfang Lin and
                  Shengnan Guo and
                  Huaiyu Wan},
  title        = {Spatial-Temporal Synchronous Graph Convolutional Networks: {A} New
                  Framework for Spatial-Temporal Network Data Forecasting},
  booktitle    = {The Thirty-Fourth {AAAI} Conference on Artificial Intelligence, {AAAI}
                  2020, The Thirty-Second Innovative Applications of Artificial Intelligence
                  Conference, {IAAI} 2020, The Tenth {AAAI} Symposium on Educational
                  Advances in Artificial Intelligence, {EAAI} 2020, New York, NY, USA,
                  February 7-12, 2020},
  pages        = {914--921},
  publisher    = {{AAAI} Press},
  year         = {2020},
  url          = {https://doi.org/10.1609/aaai.v34i01.5438},
  doi          = {10.1609/AAAI.V34I01.5438},
  bibsource    = {dblp computer science bibliography, https://dblp.org}
}

@inproceedings{10.1145/3209978.3210006,
author = {Lai, Guokun and Chang, Wei-Cheng and Yang, Yiming and Liu, Hanxiao},
title = {Modeling Long- and Short-Term Temporal Patterns with Deep Neural Networks},
year = {2018},
isbn = {9781450356572},
publisher = {Association for Computing Machinery},
address = {New York, NY, USA},
url = {https://doi.org/10.1145/3209978.3210006},
doi = {10.1145/3209978.3210006},
booktitle = {The 41st International ACM SIGIR Conference on Research \& Development in Information Retrieval},
pages = {95–104},
numpages = {10},
location = {Ann Arbor, MI, USA},
series = {SIGIR '18}
}

@article{JMLR:v25:22-0493,
  author  = {Pieter Gijsbers and Marcos L. P. Bueno and Stefan Coors and Erin LeDell and S{{\'e}}bastien Poirier and Janek Thomas and Bernd Bischl and Joaquin Vanschoren},
  title   = {AMLB: an AutoML Benchmark},
  journal = {Journal of Machine Learning Research},
  year    = {2024},
  volume  = {25},
  number  = {101},
  pages   = {1--65},
  url     = {http://jmlr.org/papers/v25/22-0493.html}
}

@inproceedings{godahewa2021monash,
  title     = {Monash Time Series Forecasting Archive},
  author    = {Godahewa, Rakshitha and Bergmeir, Christoph and Webb, Geoffrey I. and Hyndman, Rob J. and Montero-Manso, Pablo},
  booktitle = {Proceedings of the Neural Information Processing Systems Track on Datasets and Benchmarks},
  year      = {2021},
  url       = {https://datasets-benchmarks-proceedings.neurips.cc/paper/2021/hash/eddea82ad2755b24c4e168c5fc2ebd40-Abstract-round2.html}
}

@misc{karpathy2026autoresearch,
  author       = {Andrej Karpathy},
  title        = {autoresearch},
  year         = {2026},
  howpublished = {\url{https://github.com/karpathy/autoresearch}},
}

@misc{xu2026autolabfrontiermodelssolve,
      title={AutoLab: Can Frontier Models Solve Long-Horizon Auto Research and Engineering Tasks?}, 
      author={Zhangchen Xu and Junda Chen and Yue Huang and Dongfu Jiang and Jiefeng Chen and Hang Hua and Zijian Wu and Zheyuan Liu and Zexue He and Lichi Li and Shizhe Diao and Jiaxin Pei and Jinsung Yoon and Hao Zhang and Mengdi Wang and Radha Poovendran and Misha Sra and Alex Pentland and Zichen Chen},
      year={2026},
      eprint={2606.05080},
      archivePrefix={arXiv},
      primaryClass={cs.AI},
      url={https://arxiv.org/abs/2606.05080}, 
}

@misc{gao2026autoscientistsselforganizingagentteams,
      title={AutoScientists: Self-Organizing Agent Teams for Long-Running Scientific Experimentation}, 
      author={Shanghua Gao and Ada Fang and Marinka Zitnik},
      year={2026},
      eprint={2605.28655},
      archivePrefix={arXiv},
      primaryClass={cs.AI},
      url={https://arxiv.org/abs/2605.28655}, 
}

@misc{xiong2026autoresearchbenchbenchmarkingaiagents,
      title={AutoResearchBench: Benchmarking AI Agents on Complex Scientific Literature Discovery}, 
      author={Lei Xiong and Kun Luo and Ziyi Xia and Wenbo Zhang and Jin-Ge Yao and Zheng Liu and Jingying Shao and Jianlyu Chen and Hongjin Qian and Xi Yang and Qian Yu and Hao Li and Chen Yue and Xiaan Du and Yuyang Wang and Yesheng Liu and Haiyu Xu and Zhicheng Dou},
      year={2026},
      eprint={2604.25256},
      archivePrefix={arXiv},
      primaryClass={cs.AI},
      url={https://arxiv.org/abs/2604.25256}, 
}

@misc{tie2026autoresearchaiaipoweredresearch,
      title={AutoResearch AI: Towards AI-Powered Research Automation for Scientific Discovery}, 
      author={Guiyao Tie and Jiawen Shi and Dingjie Song and Yixiao Huang and Ziji Sheng and Xueyang Zhou and Daizong Liu and Pan Zhou and Yongchao Chen and Ran Xu and Lifang He and Qingsong Wen and Manling Li and Cong Lu and Shuai Li and Pengtao Xie and Yixuan Yuan and Rui Meng and Lei Xing and Lichao Sun and Caiming Xiong and Philip S. Yu and Jianfeng Gao},
      year={2026},
      eprint={2605.23204},
      archivePrefix={arXiv},
      primaryClass={cs.AI},
      url={https://arxiv.org/abs/2605.23204}, 
}

@misc{zheng2025lifelongagentbenchevaluatingllmagents,
  title={LifelongAgentBench: Evaluating LLM Agents as Lifelong Learners},
  author={Junhao Zheng and Xidi Cai and Qiuke Li and Duzhen Zhang and ZhongZhi Li and Yingying Zhang and Le Song and Qianli Ma},
  year={2025},
  eprint={2505.11942},
  archivePrefix={arXiv},
  primaryClass={cs.AI},
  url={https://arxiv.org/abs/2505.11942},
}

@inproceedings{zhao2024expel,
  title     = {{ExpeL: LLM Agents Are Experiential Learners}},
  author    = {Zhao, Andrew and Huang, Daniel and Xu, Quentin and Lin, Matthieu and Liu, Yong-Jin and Huang, Gao},
  booktitle = {Proceedings of the AAAI Conference on Artificial Intelligence},
  volume    = {38},
  number    = {17},
  pages     = {19632--19642},
  year      = {2024},
  doi       = {10.1609/aaai.v38i17.29936},
  url       = {https://ojs.aaai.org/index.php/AAAI/article/view/29936}
}

@inproceedings{3692070.3692884,
author = {Huang, Qian and Vora, Jian and Liang, Percy and Leskovec, Jure},
title = {MLAgentBench: evaluating language agents on machine learning experimentation},
year = {2024},
publisher = {JMLR.org},
booktitle = {Proceedings of the 41st International Conference on Machine Learning},
articleno = {814},
numpages = {39},
location = {Vienna, Austria},
series = {ICML'24}
}

@article{lu2026endtoend,
  author  = {Lu, Chris and Lu, Cong and Lange, Robert Tjarko and Yamada, Yutaro and Hu, Shengran and Foerster, Jakob and Ha, David and Clune, Jeff},
  title   = {Towards End-to-End Automation of AI Research},
  journal = {Nature},
  volume  = {651},
  pages   = {914--919},
  year    = {2026},
  doi     = {10.1038/s41586-026-10265-5},
  url     = {https://doi.org/10.1038/s41586-026-10265-5}
}

\newpage

\appendix
\section{Detailed Setting Descriptions} \label{app:setting-details}
In the following appendix section, we provide details about each stream in terms of design and configuration. 

\subsection{Dataset Anonymization}
To prevent agents from recognizing popular benchmark names such as ETTh1 or PEMS07 and immediately applying a known solution, we give them neutral task and dataset identifiers during their evaluation. This prevents inflation of first-solution quality and apparent search efficiency without any need for cross-task learning. 

\subsection{Synthetic AutoML}
We design the synthetic AutoML suite so that it isolates transfer across prediction objectives over a shared data source. This is inspired by companies having different modeling objectives using the same client database. The setting contains three task families each with five classification tasks generated from the same feature matrix, but with different target rules. We give the agent the knowledge that all tasks share the same data, but do not specify what is reusable across tasks.

\paragraph{Reported configuration.}
For each task family, the data generator creates one \(6000\times150\) feature matrix and divides its rows, without overlap, into 2,000 training, 2,000 development, and 2,000 private-test examples. The generators are deterministic for the reported configuration. The feature matrix, task order, splits, and labels are identical for stateful and stateless runs.

\textbf{Shared latent.}
In the shared latent task family, observed features are noisy mixtures of latent factors. Each target depends on a different subset of a shared causal pool. The ideal transfer is if the agent can use the earlier tasks to identify useful representations in the presence of correlated feature and high-variance nuisance factors.

The data generator samples 24 standard-normal latent factors. The six potentially causal factors are scaled by 0.7 and the remaining nuisance factors by 1.6 before projection through a \(24\times150\) standard-normal matrix; standard-normal feature noise is then added. For generated task index \(t\), let \(p=(t-1)\bmod6\), \(s=(t+1)\bmod6\), and \(h=(t+2)\bmod6\). The score is \(0.95z_p-0.75z_s+0.45\tanh(1.2z_5)+0.32z_h z_p+0.22\sin(z_s z_5)+\epsilon\), where \(\epsilon\sim\mathcal{N}(0,0.55^2)\).

\textbf{Shared sparse.}
In the shared sparse task family, each target depends on a changing subset of variables sampled from a small shared relevant pool. The ideal transfer is if the agent can find an efficient procedure for feature screening, regularization, and validation. 

The base features are standard normal. Features \(x_{020},\ldots,x_{029}\) are noisy copies of \(x_{000},\ldots,x_{009}\), with Gaussian noise of standard deviation 0.2. Each task selects three features without replacement from the shared pool \(x_{000},\ldots,x_{007}\) and computes \(1.7x_a-1.4x_b+0.9x_c^2+\epsilon\), where \(\epsilon\sim\mathcal{N}(0,0.35^2)\). 

\textbf{Shared regime.}
In the shared regime task family, every task contains the same regime variable. However, its effect changes depending on the target. The ideal transfer is if the agent can find earlier tasks that motivate testing for regime-dependent interactions. Each new task would then require the agent to re-estimate the interaction rather than reuse on earlier rule.

Features \(x_{000},\ldots,x_{148}\) are standard normal, and the visible feature \(x_{149}\) is the indicator \(r=\mathbb{1}[x_{000}+0.6x_{001}-0.25x_{002}>0]\). Odd tasks use \(1.7x_{003}-x_{004}+0.5x_{005}\) when \(r=0\) and \(-1.2x_{003}+2.2x_{006}-0.6x_{007}\) when \(r=1\). Even tasks use \(1.1x_{004}+0.8x_{005}^2\) when \(r=0\) and \(-1.8x_{005}+1.3x_{008}+0.4x_{009}\) when \(r=1\). Gaussian noise with standard deviation 0.35 is added. Thus, the regime definition transfers across tasks while its conditional prediction rule can change.

All three streams use a fixed five-task order. The agent receives labeled training and development data and submits development predictions, a serialized model, and an inference manifest. The private evaluator loads the model to predict test features that are unavailable to the agent. Binary tasks use private-test accuracy and multiclass tasks use private-test weighted F1.

\subsection{Symbolic Regression}
The symbolic regression setting tests whether agent are able to transfer scientific search strategy. It contains five task in a canonical-coordinates stream where the equations first appear compliecated but become simpler after an appropriate change of coordinates. Relevant transformations include centering an input, rescaling variables, and combining multiple variables into a radial coordinate. The variables, parameters, and final equation change across tasks, so the agent cannot reuse a previously discovered expression.

\paragraph{Reported configuration.}
Reported results use the canonical-coordinates family and the fixed task order shown in Table~\ref{tab:symbolic-task-construction}. Each task contains 384 training and 384 development rows, 2,000 private interpolation rows, and 2,000 private extrapolation rows. The number of generic features is 16. The data generator also randomly places the relevant variables among these columns and samples every irrelevant feature independently from \(\mathcal{U}[-2,2]\).

Let \(v_j\) denote the \(j\)-th relevant variable and \(u_j=(v_j-b_j)/s_j\) its centered and scaled value. Unless a smaller range is specified by the template, coefficient magnitudes are sampled uniformly from \([0.55,2.75]\) with a random sign, centers from \([-1.25,1.25]\), and scales from \([0.65,2.5]\). The modulation coefficient in Tasks 2 and 3 uses magnitude \([0.2,0.8]\), and the cubic coefficient in Task 4 uses \([0.15,0.65]\). Values are rounded to three decimals and stored with the sampled relevant-feature indices in a private ground-truth manifest. The resulting equation templates are:

\begin{table}[H]
\centering
\caption{Construction of the canonical-coordinates symbolic-regression stream. ``Rel./all'' gives the number of relevant and total features.}
\label{tab:symbolic-task-construction}
\begin{scriptsize}
\setlength{\tabcolsep}{3.2pt}
\renewcommand{\arraystretch}{1.12}
\begin{tabular}{@{}c>{\raggedright\arraybackslash}p{0.68\textwidth}cc@{}}
\toprule
Task & Equation template & Rel./all & Control \\
\midrule
1 & \(c+a u_1^2+q u_1\sin(u_2)+r\cos(v_3+u_1u_2)\) & 3/9 & No \\
2 & \(c+a\exp(-u_1^2)(1+q\cos(u_2))+r\sin(v_3u_1)\) & 3/10 & No \\
3 & \(c+a\exp[-(u_1^2+u_2^2+u_3^2)](1+q\cos(v_4))\) & 4/12 & No \\
4 & \(c+a u_1^2+k u_1^3+q\sin(u_2)+r\cos(v_3v_4)\) & 4/14 & Yes \\
5 & \(c+a\cos(\sqrt{u_1^2+u_2^2+u_3^2})+q\sin(v_4v_5)\) & 5/16 & No \\
\bottomrule
\end{tabular}
\end{scriptsize}
\end{table}

Training, development, and interpolation inputs use standardized radius at most 1.5 for the centered variables. Extrapolation samples use radii between 1.5 and 3.5 for Tasks 1, 2, and 4, and between 1.5 and 4.5 for Tasks 3 and 5; the additional signed variables remain within their training range. Training and development targets receive independent Gaussian noise with standard deviation equal to 1\% of the clean training-target standard deviation. Both private target sets are noise-free.

The agent submits an \texttt{expression.json} file containing one expression. The restricted evaluator supports \(+,-,\times,/\), \texttt{**}, finite constants, and \texttt{sin}, \texttt{cos}, \texttt{exp}, \texttt{log}, \texttt{sqrt}, and \texttt{abs}; it parses and evaluates the expression without an LLM judge. Development \(R^2\) is visible during search. We use the arithmetic mean of interpolation and extrapolation \(R^2\). We also record each component \(R^2\), normalized MSE, expression-tree complexity, and whether symbolic simplification recovers the generating structure. Task 4 is asymmetric because of its cubic term and tests whether an agent verifies rather than blindly reuses an even-symmetry assumption.

\subsection{Time-Series Forecasting}
In the time series forecasting setting, we design three task family streams with various degree of shared structure. All tasks require the agent to submit an executable forecaster from the data we provide. 

\textbf{ETT cross-resolution.}
This stream contains, in order, ETTm1, ETTh1, ETTh2, and ETTm2 \citep{zhou2021aaai-informer}. ETTh1 and ETTh2 use hourly resolution, while ETTm1 and ETTm2 use 15-minute resolution. The datasets measure the same type of electricity-transformer system, but differ in station and sampling frequency. This tests whether an agent transfers insights about normalization, lag construction, and periodicity while adapting context lengths and seasonal periods to the new resolution.

\textbf{PEMS traffic networks.}
This stream contains, in order, PEMS03, PEMS04, PEMS07, and PEMS08 \citep{DBLP:conf/aaai/SongLGW20}. Each dataset represents traffic over a different sensor network and time period. The components that are transferrable are for handling correlated channels, daily and weekly seasonality, and chronological validation. The different network prevents the agent from simply copying a fitted model. 

\textbf{Energy systems.}
This stream contains, in order, Electricity Hourly, Australian Electricity Demand, Solar 10 Minutes, ETTh1, and ETTh2 \citep{10.1145/3209978.3210006,godahewa2021monash,zhou2021aaai-informer}. It is less homogeneous than the ETT stream. However, it tests for forecasting practices such as periodicity checking, seasonal baselines, scaling channels, and others.  

All forecasting tasks are evaluated by RMSE on private chronological test splits.

The three streams looks at different forms of transfer. ETT tests adaptation within a closely related system, PEMS tests reuse across related sensor networks, and the energy stream tests whether a forecasting workflow survives broader domain variation.

\subsection{Realistic AutoML}
The tabular AutoML setting contains two subsets of existing evaluation benchmarks. We use a subset of SELA's evaluation, texttt{credit-g}, \texttt{kick}, \texttt{Click\_prediction\_small}, \texttt{software-defects}, and \texttt{smoker-status} \citep{chi2024selatreesearchenhancedllm}, and use a three task subset from MLE-Bench, \texttt{spaceship-titanic}, \texttt{tabular-playground-series-may-2022}, and \texttt{tabular-playground-series-dec-2021} \citep{chan2025mlebench}.

In contrast to the other task families, the tasks don't share a data generator or a guaranteed predictive structure. For the SELA evaluation subset, the agent controls preprocessing, feature engineering, validation, and model selection before submitting an executable predictor, which we evaluate using F1 on a private test set. The MLE-Bench subset retains its competition-style prediction-submission protocol: \texttt{spaceship-titanic} and the December 2021 task use accuracy, while the May 2022 task uses ROC-AUC.

\section{Complete Results} \label{app:complete-results}

Table~\ref{tab:model-results-aggregate-appendix} summarizes results by model, Table~\ref{tab:task-nature-aggregate} groups results by task nature, and Table~\ref{tab:main-results-complete} reports the corresponding task family-level results underlying the comparisons in Section~\ref{sec:results}.

\begin{table}[H]
\centering
\caption{Model-level results over transfer-eligible tasks. Gains are normalized reward points and are reported as mean [median]. Aggregates use equal task-family weights. S/R denotes stateful/reset.}
\label{tab:model-results-aggregate-appendix}
\begin{scriptsize}
\setlength{\tabcolsep}{1.8pt}
\renewcommand{\arraystretch}{1.08}
\begin{tabular}{@{}
>{\raggedright\arraybackslash}p{0.175\linewidth}
>{\centering\arraybackslash}p{0.040\linewidth}
>{\centering\arraybackslash}p{0.105\linewidth}
>{\centering\arraybackslash}p{0.105\linewidth}
>{\centering\arraybackslash}p{0.105\linewidth}
>{\centering\arraybackslash}p{0.075\linewidth}
>{\centering\arraybackslash}p{0.075\linewidth}
>{\centering\arraybackslash}p{0.075\linewidth}
>{\centering\arraybackslash}p{0.075\linewidth}
@{}}
\toprule
Model & Tasks & \makecell{Outcome\\gain} & \makecell{Init.\\gain} & \makecell{Time-conv.\\gain} & \makecell{Outcome\\win rate} & \makecell{Negative\\transfer} & \makecell{Valid\\S/R} & \makecell{Failed\\S/R} \\
\midrule
Opus 5 & 32 & -0.9 [+0.0] & +1.3 [+0.5] & +7.5 [+10.3] & 34.3\% & 65.7\% & 147/149 & 13/16 \\
GPT-5.6-Sol & 32 & -0.8 [+0.2] & -1.0 [+1.2] & +6.1 [+7.4] & 59.3\% & 40.7\% & 199/397 & 18/7 \\
Kimi K3 & 32 & +0.4 [+2.0] & +1.8 [+2.0] & +21.3 [+27.5] & 62.0\% & 38.0\% & 108/109 & 2/14 \\
Sonnet 4.6 & 32 & +7.6 [+1.3] & +7.0 [+7.0] & +20.9 [+20.6] & 66.7\% & 33.3\% & 180/160 & 15/71 \\
GPT-5.4 & 32 & +0.2 [-0.2] & +3.0 [+3.1] & +3.0 [+2.0] & 51.9\% & 48.1\% & 153/200 & 20/23 \\
\bottomrule
\end{tabular}
\end{scriptsize}
\end{table}

\begin{table}[H]
\centering
\caption{Transfer by task nature. Classification combines the synthetic and realistic AutoML streams, regression combines the time-series streams, and symbolic regression is reported separately. Gains are normalized reward points reported as mean [median] over model--family means, with each model--family pair equally weighted. Task counts exclude the first task of each stream and are reported per model. Win and negative-transfer rates pool transfer-task outcomes across the five models.}
\label{tab:task-nature-aggregate}
\begin{scriptsize}
\setlength{\tabcolsep}{3.2pt}
\renewcommand{\arraystretch}{1.08}
\begin{tabular}{@{}lccrrrcc@{}}
\toprule
Task nature & Families & Tasks & \makecell{Outcome\\gain} & \makecell{Init.\\gain} & \makecell{Time-conv.\\gain} & \makecell{Outcome\\win rate} & \makecell{Negative\\transfer} \\
\midrule
Classification & 5 & 18 & +0.8 [+0.4] & +3.4 [+2.0] & +16.4 [+18.4] & 60.0\% & 40.0\% \\
Regression & 3 & 10 & +0.8 [-0.1] & +3.4 [+1.7] & +3.7 [+7.4] & 50.0\% & 50.0\% \\
Symbolic reg. & 1 & 4 & +5.2 [-2.9] & -5.6 [+1.6] & +12.6 [+5.0] & 55.0\% & 45.0\% \\
\bottomrule
\end{tabular}
\end{scriptsize}
\end{table}

{\scriptsize
\setlength{\tabcolsep}{1.6pt}
\renewcommand{\arraystretch}{1.02}
\begin{longtable}{@{}
>{\raggedright\arraybackslash}p{0.105\linewidth}
>{\raggedright\arraybackslash}p{0.130\linewidth}
>{\centering\arraybackslash}p{0.035\linewidth}
>{\centering\arraybackslash}p{0.105\linewidth}
>{\centering\arraybackslash}p{0.105\linewidth}
>{\centering\arraybackslash}p{0.105\linewidth}
>{\centering\arraybackslash}p{0.065\linewidth}
>{\centering\arraybackslash}p{0.065\linewidth}
>{\centering\arraybackslash}p{0.070\linewidth}
>{\centering\arraybackslash}p{0.070\linewidth}
@{}}
\caption{Complete stateful-versus-reset results over transfer-eligible tasks. Gains are normalized reward points and are reported as mean [median]. S/R denotes stateful/reset, and TS denotes time series. Overall rows use equal family weights; family rows summarize tasks. Task 1 is excluded.}
\label{tab:main-results-complete}\\
\toprule
Setting & Task family & Tasks & \makecell{Outcome\\gain} & \makecell{Init.\\gain} & \makecell{Time-conv.\\gain} & \makecell{Outcome\\win rate} & \makecell{Negative\\transfer} & \makecell{Valid\\S/R} & \makecell{Failed\\S/R} \\
\midrule
\endfirsthead
\multicolumn{10}{c}{\tablename~\thetable{} -- continued} \\
\toprule
Setting & Task family & Tasks & \makecell{Outcome\\gain} & \makecell{Init.\\gain} & \makecell{Time-conv.\\gain} & \makecell{Outcome\\win rate} & \makecell{Negative\\transfer} & \makecell{Valid\\S/R} & \makecell{Failed\\S/R} \\
\midrule
\endhead
\midrule
\multicolumn{10}{r}{Continued on next page} \\
\endfoot
\bottomrule
\endlastfoot
\multicolumn{10}{@{}l}{\textbf{Claude Code / Opus 5}} \\*
\bfseries All settings & \bfseries Overall & 32 & -0.9 [+0.0] & +1.3 [+0.5] & +7.5 [+10.3] & 34.3\% & 65.7\% & 147/149 & 13/16 \\
Synth. AutoML & Shared latent & 4 & -0.1 [-0.1] & +0.1 [+0.2] & +3.1 [+4.2] & 25.0\% & 75.0\% & 13/13 & 0/0 \\
Synth. AutoML & Shared regime & 4 & +0.5 [+0.7] & +12.7 [+9.6] & +24.0 [+17.9] & 75.0\% & 25.0\% & 23/20 & 2/2 \\
Synth. AutoML & Shared sparse & 4 & -3.5 [-3.6] & -1.2 [-1.2] & +24.5 [+22.1] & 0.0\% & 100.0\% & 16/16 & 0/1 \\
Symbolic reg. & Canonical coord. & 4 & +1.3 [+0.0] & +3.7 [-0.1] & +29.0 [+29.0] & 25.0\% & 75.0\% & 26/20 & 8/0 \\
TS forecasting & Energy systems & 4 & -6.6 [-0.3] & -7.2 [-0.2] & +10.3 [+10.1] & 25.0\% & 75.0\% & 23/16 & 0/1 \\
TS forecasting & ETT cross-res. & 3 & +0.0 [+0.2] & +0.5 [+0.2] & -6.1 [-5.6] & 66.7\% & 33.3\% & 13/11 & 3/5 \\
TS forecasting & Traffic network & 3 & +0.5 [+0.8] & +1.7 [+1.7] & +8.6 [+12.8] & 66.7\% & 33.3\% & 10/22 & 0/3 \\
Real. AutoML & MLE-Bench & 2 & -1.5 [-1.5] & -1.3 [-1.3] & -44.5 [-44.5] & 0.0\% & 100.0\% & 4/15 & 0/1 \\
Real. AutoML & SELA eval. subset & 4 & +1.4 [-0.8] & +2.4 [+1.0] & +18.4 [+26.6] & 25.0\% & 75.0\% & 19/16 & 0/3 \\
\addlinespace[3pt]
\multicolumn{10}{@{}l}{\textbf{Codex / GPT-5.6-Sol}} \\*
\bfseries All settings & \bfseries Overall & 32 & -0.8 [+0.2] & -1.0 [+1.2] & +6.1 [+7.4] & 59.3\% & 40.7\% & 199/397 & 18/7 \\
Synth. AutoML & Shared latent & 4 & -0.2 [-0.1] & +0.7 [+0.7] & +9.9 [+9.5] & 0.0\% & 100.0\% & 19/21 & 0/0 \\
Synth. AutoML & Shared regime & 4 & +0.9 [+0.9] & +11.6 [+6.7] & +4.7 [-2.5] & 100.0\% & 0.0\% & 16/82 & 0/2 \\
Synth. AutoML & Shared sparse & 4 & +0.2 [+0.1] & +7.3 [+8.6] & -0.9 [-2.0] & 50.0\% & 50.0\% & 17/85 & 0/1 \\
Symbolic reg. & Canonical coord. & 4 & -2.9 [+0.0] & -34.5 [-25.7] & +5.0 [-2.5] & 50.0\% & 50.0\% & 24/26 & 9/1 \\
TS forecasting & Energy systems & 4 & -3.9 [-0.9] & +1.6 [+1.6] & +8.0 [-4.6] & 25.0\% & 75.0\% & 32/82 & 0/0 \\
TS forecasting & ETT cross-res. & 3 & +0.4 [+0.1] & +1.2 [+0.5] & +7.4 [+9.3] & 100.0\% & 0.0\% & 18/13 & 0/0 \\
TS forecasting & Traffic network & 3 & -2.2 [-3.3] & +3.8 [+3.9] & +17.0 [+3.8] & 33.3\% & 66.7\% & 35/15 & 0/0 \\
Real. AutoML & MLE-Bench & 2 & +0.2 [+0.2] & +0.3 [+0.3] & +15.1 [+15.1] & 100.0\% & 0.0\% & 25/20 & 2/1 \\
Real. AutoML & SELA eval. subset & 4 & +0.4 [+0.6] & -0.6 [-0.9] & -11.5 [-12.9] & 75.0\% & 25.0\% & 13/53 & 7/2 \\
\addlinespace[3pt]
\multicolumn{10}{@{}l}{\textbf{OpenCode / Kimi K3}} \\*
\bfseries All settings & \bfseries Overall & 32 & +0.4 [+2.0] & +1.8 [+2.0] & +21.3 [+27.5] & 62.0\% & 38.0\% & 108/109 & 2/14 \\
Synth. AutoML & Shared latent & 4 & +0.9 [+0.3] & +4.0 [+2.5] & +32.2 [+29.3] & 75.0\% & 25.0\% & 11/13 & 0/0 \\
Synth. AutoML & Shared regime & 4 & +2.1 [+2.2] & +2.0 [+2.0] & +48.2 [+46.4] & 100.0\% & 0.0\% & 16/11 & 0/2 \\
Synth. AutoML & Shared sparse & 4 & +4.3 [+4.6] & +4.3 [+4.8] & +34.2 [+34.4] & 100.0\% & 0.0\% & 9/16 & 0/1 \\
Symbolic reg. & Canonical coord. & 4 & -10.9 [-1.0] & +1.6 [-0.8] & +33.9 [+38.8] & 25.0\% & 75.0\% & 24/8 & 0/2 \\
TS forecasting & Energy systems & 4 & +3.5 [-0.1] & +3.5 [+0.0] & -13.7 [-9.9] & 25.0\% & 75.0\% & 9/9 & 1/5 \\
TS forecasting & ETT cross-res. & 3 & -0.4 [-0.5] & -1.1 [-1.6] & +2.4 [+5.3] & 33.3\% & 66.7\% & 7/7 & 0/1 \\
TS forecasting & Traffic network & 3 & +2.6 [+2.2] & +2.8 [+2.3] & +8.1 [+6.3] & 100.0\% & 0.0\% & 3/5 & 0/0 \\
Real. AutoML & MLE-Bench & 2 & -0.8 [-0.8] & -0.4 [-0.4] & +27.5 [+27.5] & 0.0\% & 100.0\% & 15/16 & 0/2 \\
Real. AutoML & SELA eval. subset & 4 & +2.0 [+2.2] & -0.9 [-0.1] & +18.9 [+22.4] & 100.0\% & 0.0\% & 14/24 & 1/1 \\
\addlinespace[3pt]
\multicolumn{10}{@{}l}{\textbf{Claude Code / Sonnet 4.6}} \\*
\bfseries All settings & \bfseries Overall & 32 & +7.6 [+1.3] & +7.0 [+7.0] & +20.9 [+20.6] & 66.7\% & 33.3\% & 180/160 & 15/71 \\
Synth. AutoML & Shared latent & 4 & +5.0 [+3.8] & +7.0 [+6.9] & +27.9 [+26.1] & 75.0\% & 25.0\% & 24/26 & 1/15 \\
Synth. AutoML & Shared regime & 4 & +3.0 [+0.3] & +8.7 [+7.7] & +28.8 [+21.8] & 75.0\% & 25.0\% & 24/28 & 3/14 \\
Synth. AutoML & Shared sparse & 4 & -0.2 [+0.3] & +2.4 [+1.1] & +8.3 [+3.0] & 50.0\% & 50.0\% & 16/32 & 0/19 \\
Symbolic reg. & Canonical coord. & 4 & +41.2 [+32.1] & +11.5 [+7.9] & -2.7 [+1.0] & 100.0\% & 0.0\% & 19/15 & 1/2 \\
TS forecasting & Energy systems & 4 & +18.5 [+16.5] & +21.0 [+20.5] & +20.6 [+27.8] & 75.0\% & 25.0\% & 19/21 & 1/4 \\
TS forecasting & ETT cross-res. & 3 & -0.1 [+0.3] & -1.6 [-3.2] & +17.7 [+39.6] & 66.7\% & 33.3\% & 6/7 & 2/2 \\
TS forecasting & Traffic network & 3 & -0.1 [-0.2] & +12.9 [+1.8] & +2.6 [+11.9] & 33.3\% & 66.7\% & 10/11 & 1/3 \\
Real. AutoML & MLE-Bench & 2 & -0.3 [-0.3] & -0.6 [-0.6] & +62.4 [+62.4] & 50.0\% & 50.0\% & 11/4 & 5/7 \\
Real. AutoML & SELA eval. subset & 4 & +1.3 [+1.9] & +1.8 [+2.2] & +22.0 [+6.2] & 75.0\% & 25.0\% & 51/16 & 1/5 \\
\addlinespace[3pt]
\multicolumn{10}{@{}l}{\textbf{Codex / GPT-5.4}} \\*
\bfseries All settings & \bfseries Overall & 32 & +0.2 [-0.2] & +3.0 [+3.1] & +3.0 [+2.0] & 51.9\% & 48.1\% & 153/200 & 20/23 \\
Synth. AutoML & Shared latent & 4 & +0.5 [+0.1] & +5.2 [+2.0] & +10.3 [+12.5] & 50.0\% & 50.0\% & 24/28 & 0/4 \\
Synth. AutoML & Shared regime & 4 & -0.4 [-0.2] & +12.0 [+8.4] & +32.3 [+37.3] & 50.0\% & 50.0\% & 16/32 & 0/1 \\
Synth. AutoML & Shared sparse & 4 & +0.4 [+0.4] & +8.5 [+6.6] & +9.0 [+8.9] & 75.0\% & 25.0\% & 39/22 & 0/4 \\
Symbolic reg. & Canonical coord. & 4 & -2.9 [+1.3] & -10.2 [-0.3] & -2.3 [+8.3] & 75.0\% & 25.0\% & 21/29 & 18/9 \\
TS forecasting & Energy systems & 4 & -1.6 [-1.4] & +6.4 [+4.3] & +0.3 [+2.1] & 50.0\% & 50.0\% & 12/29 & 0/0 \\
TS forecasting & ETT cross-res. & 3 & -0.7 [-0.5] & +3.1 [+3.4] & +2.0 [-5.5] & 0.0\% & 100.0\% & 11/15 & 0/0 \\
TS forecasting & Traffic network & 3 & +1.5 [+1.4] & +1.7 [+1.3] & -29.2 [-27.3] & 66.7\% & 33.3\% & 6/10 & 0/2 \\
Real. AutoML & MLE-Bench & 2 & -0.2 [-0.2] & -0.6 [-0.6] & +1.7 [+1.7] & 0.0\% & 100.0\% & 9/17 & 2/2 \\
Real. AutoML & SELA eval. subset & 4 & +5.0 [+4.7] & +0.7 [+0.3] & +2.5 [-4.0] & 100.0\% & 0.0\% & 15/18 & 0/1 \\
\end{longtable}
}

\begin{figure}[H]
    \centering
    \includegraphics[width=0.9\textwidth]{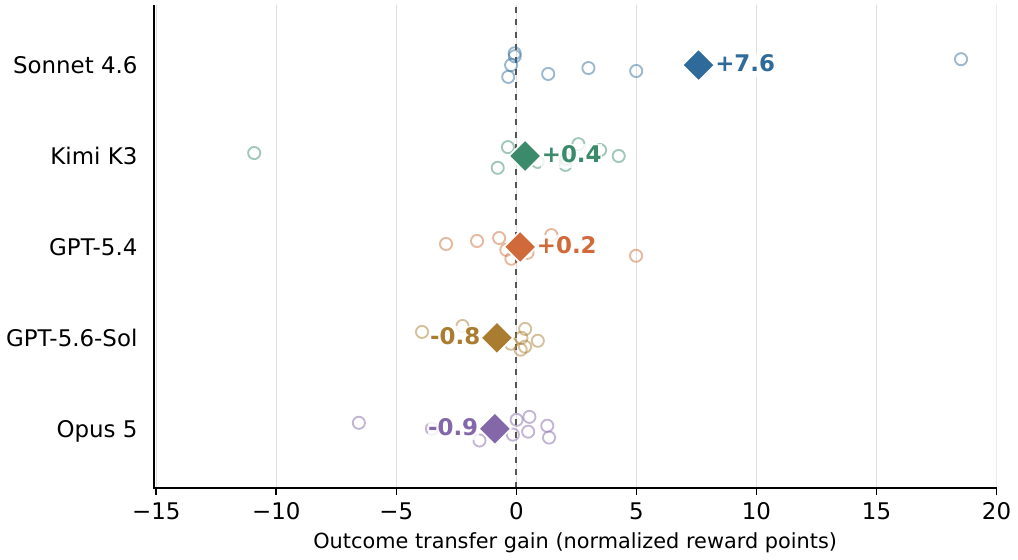}
    \caption{Outcome transfer gain measures the difference in final selected hidden-test reward between persistent and reset execution on the same task (stateful minus stateless). Results cover 32 transfer tasks per agent, excluding the first task of every stream. Values are normalized reward points (\(100\times\) reward difference), so positive values favor persistent state. Sonnet 4.6 shows the largest aggregate gain (\(+7.6\) points); GPT-5.4 and Kimi K3 remain near zero, while Opus 5 and GPT-5.6-Sol are slightly negative.}
    \label{fig:outcome_gain}
\end{figure}

\begin{figure}[H]
    \centering
    \includegraphics[width=0.9\textwidth]{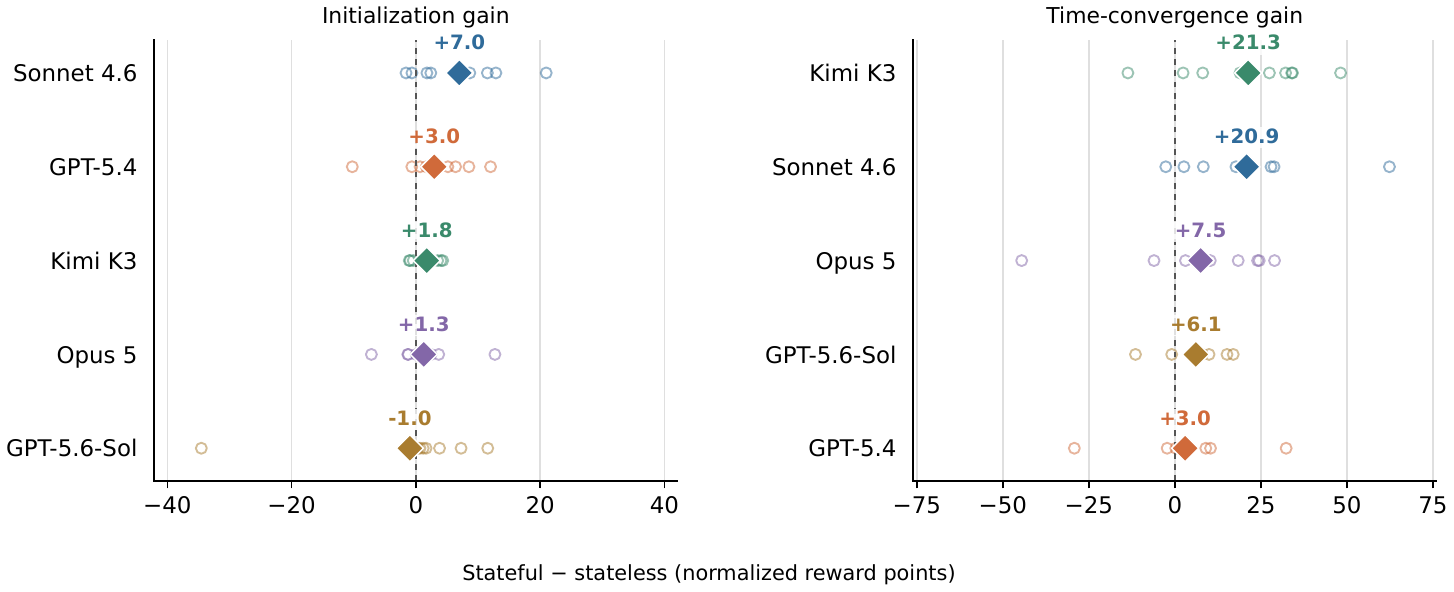}
    \caption{Initialization gain compares first valid solutions; time-convergence gain summarizes the best-so-far reward over the task budget. Open circles are task-family means, and diamonds are equal-family agent means. Gains are stateful minus stateless normalized reward points, so zero indicates no difference and positive values favor stateful execution. Mean time-convergence gain is positive for all five agents, and mean initialization gain is positive for four of five.}
    \label{fig:research_process_gain}
\end{figure}

\begin{figure}[H]
    \centering
    \includegraphics[width=\textwidth]{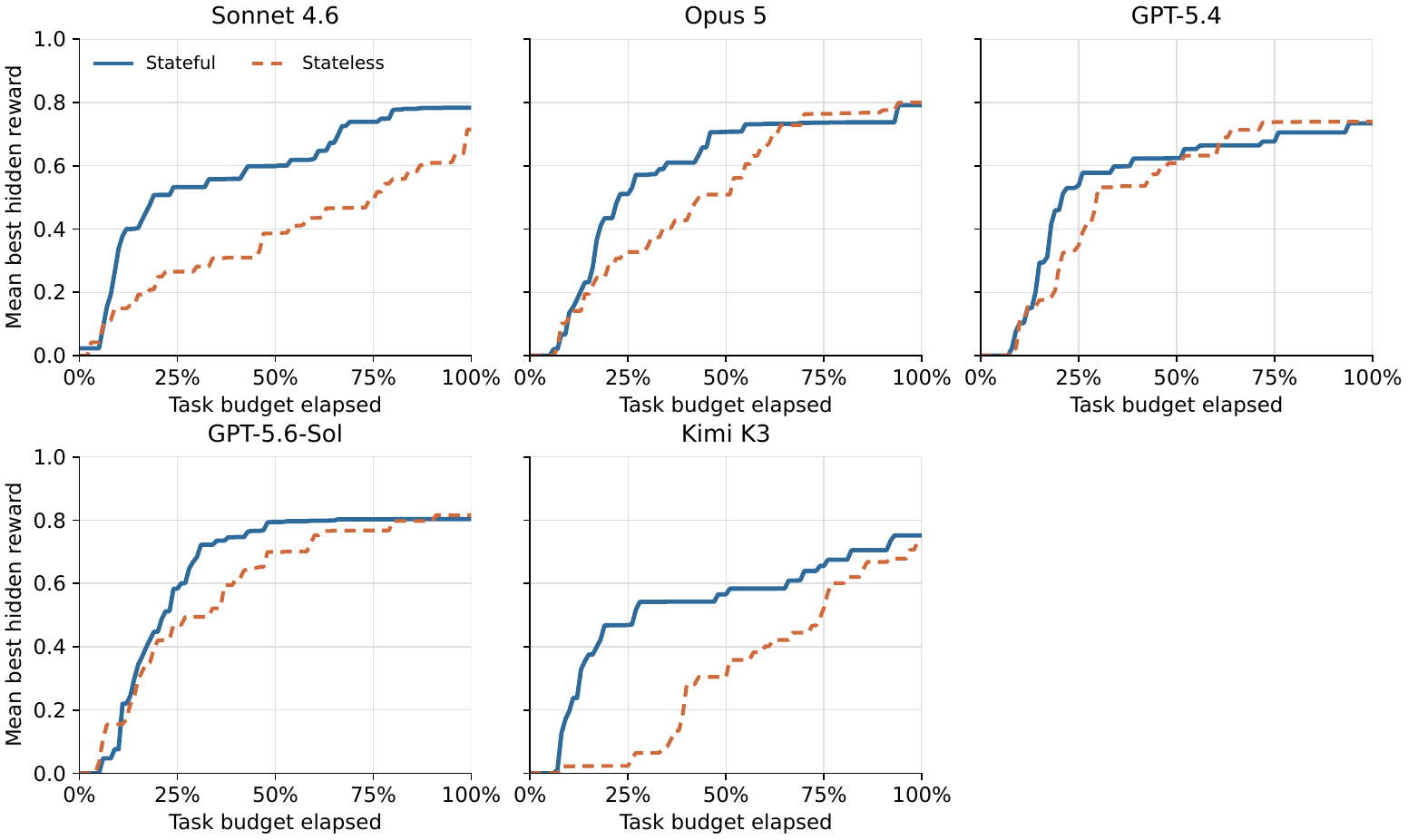}
    \caption{Curves show the mean best hidden reward attained as a function of normalized task-budget elapsed. Task-level best-so-far curves are constructed post hoc from private-evaluator scores, averaged within each family, and then averaged with equal family weight. Persistent execution generally finds useful solutions earlier, but the size and duration of the advantage vary across agents.}
    \label{fig:search_trajectories_all}
\end{figure}

\begin{figure}[H]
    \centering
    \includegraphics[width=\textwidth]{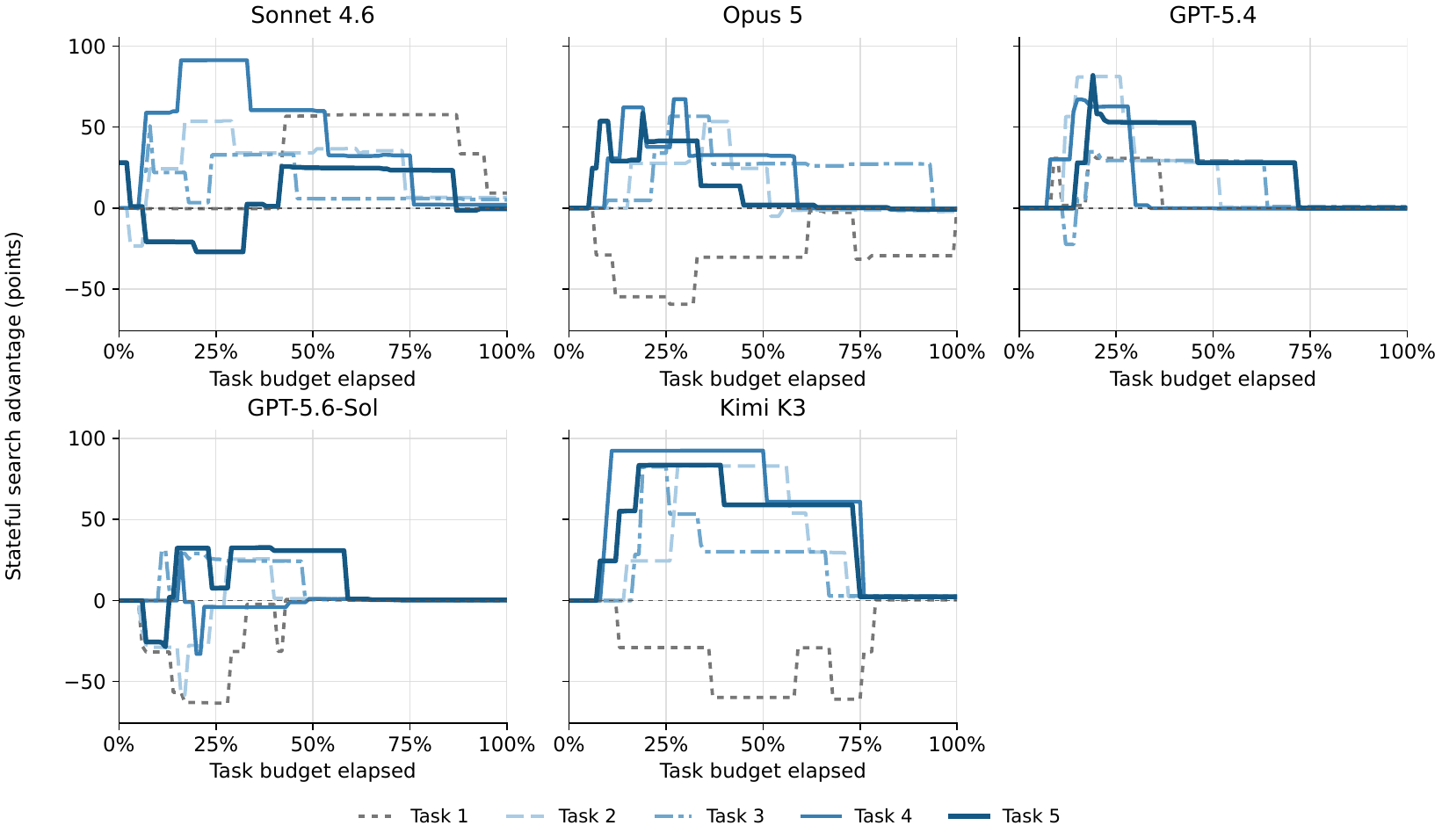}
    \caption{Task-position-resolved search advantage on the controlled synthetic streams. Each curve reports the stateful-minus-stateless best hidden reward, in normalized reward points, over the normalized task budget. Differences are computed within the same task and then averaged with equal weight across the shared-latent, shared-regime, and shared-sparse families. Later tasks exhibit a pronounced and sustained stateful advantage for several agents, although the pattern is not universal across models.}
    \label{fig:synthetic-task-position-trajectories}
\end{figure}

\begin{figure}[H]
    \centering
    \includegraphics[width=\textwidth]{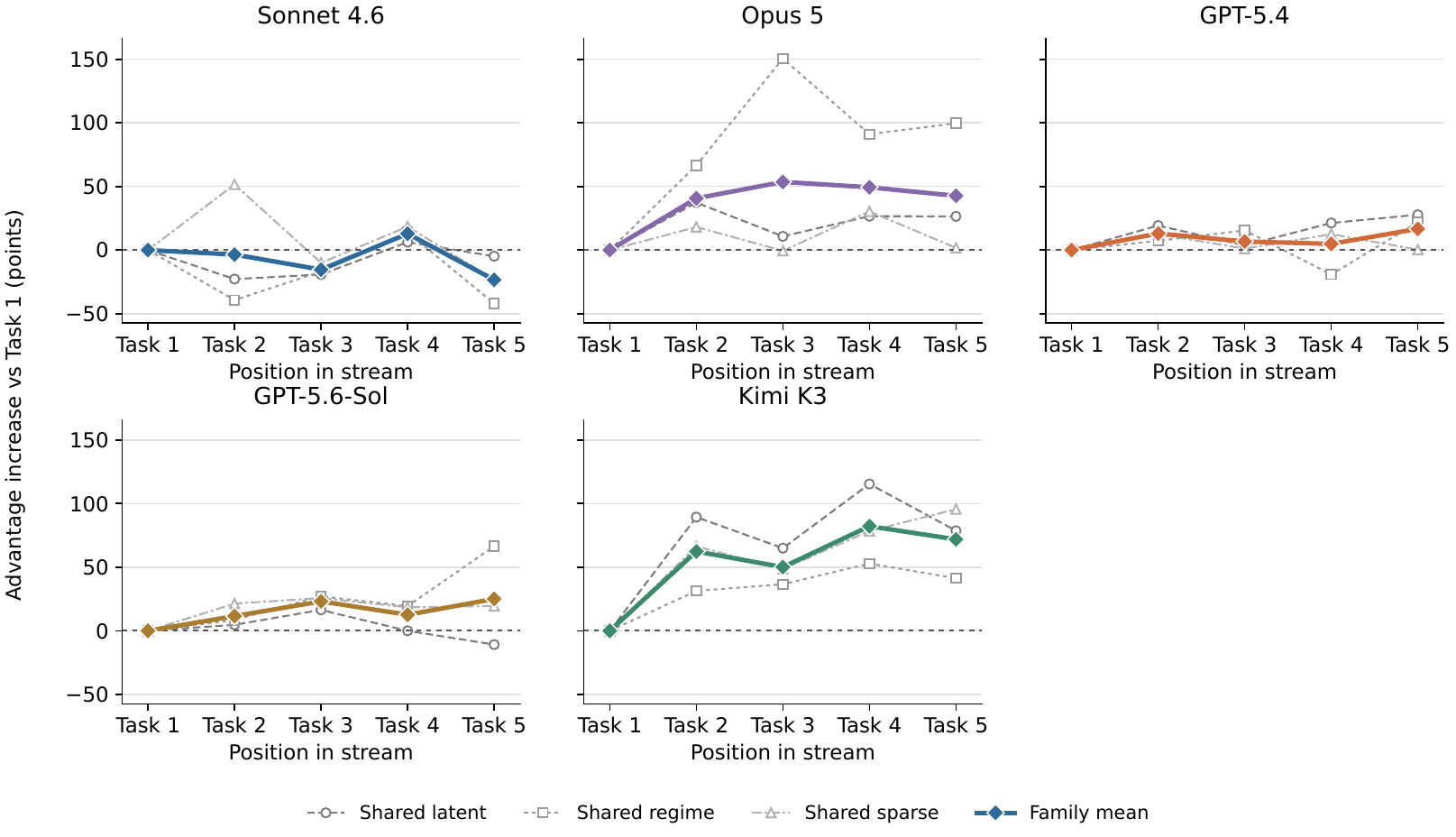}
    \caption{Growth in persistent-state search advantage across the controlled synthetic streams. For each task position, the stateful-minus-stateless reward gap is integrated over the normalized task budget and compared with the corresponding initial-task value from the same model and family. Gray lines show the shared-latent, shared-regime, and shared-sparse families; colored diamonds show equal-family means. Positive values indicate that persistent state provides a larger search advantage on later tasks.}
    \label{fig:synthetic-task-position-summary}
\end{figure}

Figures~\ref{fig:outcome_gain}, \ref{fig:research_process_gain}, and~\ref{fig:search_trajectories_all} provide complementary views of outcome, initialization, and search-trajectory performance. Figures~\ref{fig:synthetic-task-position-trajectories} and~\ref{fig:synthetic-task-position-summary} decompose the controlled synthetic results by task position.

\begin{figure}[H]
    \centering
    \includegraphics[width=\textwidth]{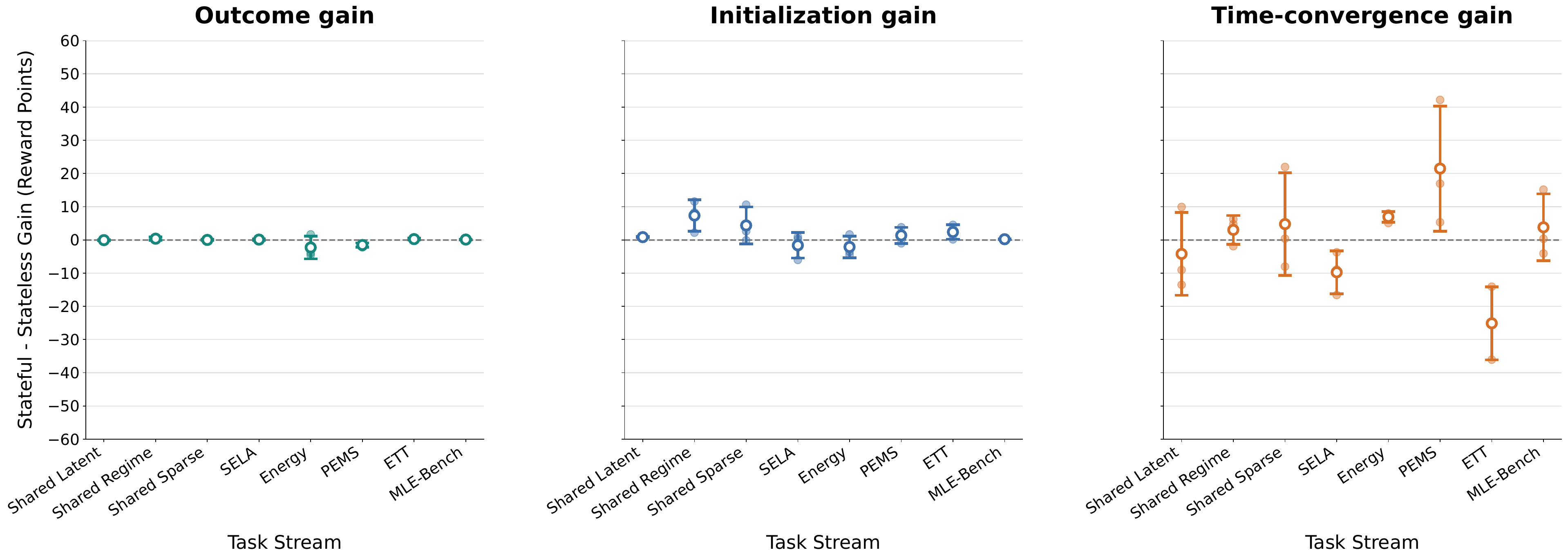}
    \caption{Run-to-run variability for GPT-5.6-Sol across three repeated stateful--reset comparisons. Dots show individual runs, circles their mean, and error bars \(\pm1\) standard deviation; positive values favor stateful execution.}
    \label{fig:stochasticity}
\end{figure}

\begin{figure}[H]
    \centering
    \includegraphics[width=\textwidth]{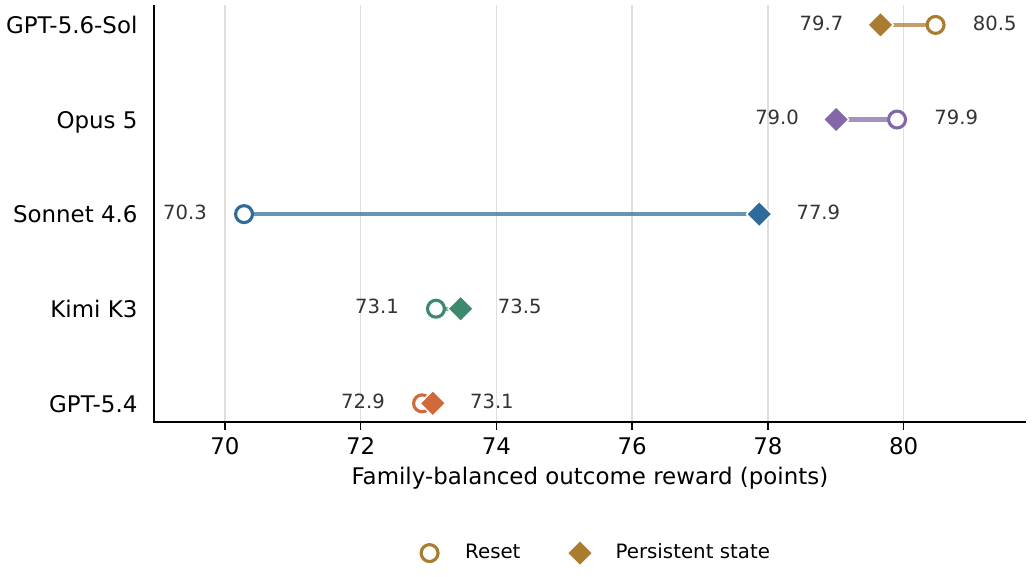}
    \caption{Absolute outcome performance under reset and persistent execution. Values are normalized reward points, averaged within each task family and then across families with equal weight. The analysis covers 32 transfer-eligible tasks per agent and excludes the first task of every stream. Opus 5 and GPT-5.6-Sol achieve the strongest reset performance, whereas Sonnet 4.6 benefits substantially from persistent state and nearly closes the absolute-performance gap.}
    \label{fig:absolute-outcome-by-agent}
\end{figure}

\begin{figure}[H]
    \centering
    \includegraphics[width=\textwidth]{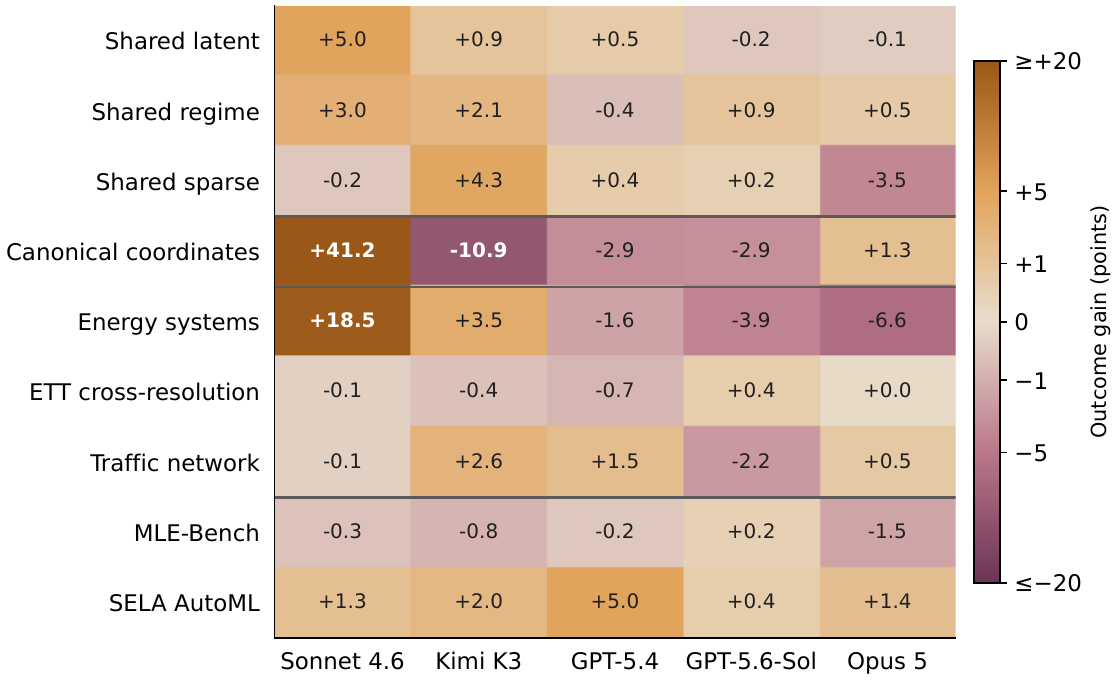}
    \caption{Outcome transfer gain decomposed by agent and task family. Each cell reports the mean persistent-minus-reset outcome reward over the transfer-eligible tasks in that family, in normalized reward points; positive values favor persistent state. Cell labels show the uncapped values, while color intensity uses a signed square-root scale capped at $\pm20$ points to preserve visibility of smaller effects. Transfer is strongly heterogeneous across both agents and recurring task structures.}
    \label{fig:outcome-gain-by-family}
\end{figure}

\subsection{Time-Convergence Robustness} \label{app:time-convergence-robustness}

We show that time-convergence does not increase by simply increasing the number of submissions. In fact, after the first valid solution, a new candidate changes the best-so-far curve only if it improves hidden reward. Instead, the metric rewards both producing a solution sooner while still being a strong one. The candidate-matched and post-first controls in Table~\ref{tab:time-convergence-robustness} deliberately remove parts of this fixed-budget estimand to identify its mechanism. They show that Sonnet retains a clear advantage after both controls, whereas the large Kimi gain and the smaller Opus and GPT-5.6-Sol gains arise primarily from producing their first valid solutions sooner.

\begin{table}[H]
\centering
\caption{Decomposition and robustness of time-convergence gains over 32 transfer tasks per agent. Time to first valid solution and valid candidate counts are family-balanced per-task means reported as stateful/reset (S/R). First-solution gain compares the hidden reward of the first valid solutions. Full-budget gain is the headline time-convergence result. The same-candidate-count control uses only the first \(\min(n_S,n_R)\) valid candidates from each paired run. The final control starts each trajectory at its own first valid solution, removing first-solution latency while preserving first-solution quality. All gains are normalized reward points; positive values favor stateful execution.}
\label{tab:time-convergence-robustness}
\begin{small}
\setlength{\tabcolsep}{2.5pt}
\renewcommand{\arraystretch}{1.08}
\begin{tabular*}{0.88\textwidth}{@{\extracolsep{\fill}}lcccccc@{}}
\toprule
Agent & \makecell{Time to first\\valid (min)\\S/R} & \makecell{Valid\\candidates\\S/R} & \makecell{First-solution\\gain} & \makecell{Full-budget\\gain} & \makecell{Gain with same\\candidate count} & \makecell{Gain excluding\\first-solution\\latency} \\
\midrule
Opus 5      & 7.7/11.1 & 4.4/4.9  & +1.3 & +7.5  & -0.5 & -0.3 \\
GPT-5.6-Sol & 6.1/8.5  & 6.7/11.8 & -1.0 & +6.1  & +0.1 & -1.1 \\
Kimi K3     & 9.5/18.7 & 3.5/3.6  & +1.8 & +21.3 & +1.7 & +1.5 \\
Sonnet 4.6  & 7.7/16.3 & 5.5/4.7  & +7.0 & +20.9 & +6.5 & +6.6 \\
GPT-5.4     & 8.4/9.2  & 4.7/6.3  & +3.0 & +3.0  & +1.2 & +0.5 \\
\bottomrule
\end{tabular*}
\end{small}
\end{table}

\paragraph{Sensitivity to the evaluation horizon.}
While time convergence represents the whole reward trajectory, it does not show whether the advantage is concentrated in early parts or persists to the deadline. Table~\ref{tab:budget-checkpoints} compares the best hidden reward achieve at four different checkpoints. Stateful execution leads for all five agents at 25\% and 50\% of the budget. The mean gap decreases from $+24.0$ points at 25\% to $+1.2$ at the deadline, where only Kimi K3 and Sonnet 4.6 remain positive. These results show that persistence provides the most consistent advantage early in the research process. 

\begin{table}[H]
\centering
\caption{Stateful-minus-reset difference in the best hidden reward achieved by each budget checkpoint. Values are normalized reward points aggregated with equal task-family weights over 32 transfer tasks per agent. Positive values favor stateful execution; the mean gives equal weight to each agent.}
\label{tab:budget-checkpoints}
\begin{scriptsize}
\setlength{\tabcolsep}{7.0pt}
\renewcommand{\arraystretch}{1.08}
\begin{tabular}{@{}lrrrr@{}}
\toprule
Agent & 25\% & 50\% & 75\% & 100\% \\
\midrule
Opus 5      & +18.3 & +19.8 & -2.8 & -0.8 \\
GPT-5.6-Sol & +11.4 &  +9.5 & +3.5 & -1.2 \\
Kimi K3     & +44.6 & +26.1 & +13.2 & +1.7 \\
Sonnet 4.6  & +26.7 & +21.3 & +24.6 & +6.9 \\
GPT-5.4     & +18.9 &  +1.6 & -6.2 & -0.5 \\
\midrule
Mean        & +24.0 & +15.7 & +6.5 & +1.2 \\
\bottomrule
\end{tabular}
\end{scriptsize}
\end{table}

\paragraph{Quality at matched candidate numbers.}
Table~\ref{tab:reward-by-candidate} compares best reward under the same number of valid candidates. Sonnet demonstrates strong first solution, and still retains strong advantage at later iterations. GPT-5.4 starts strong, but is not able to keep the difference. Coverage falls at later candidate numbers because both paired runs must have produced at least that many valid solutions. 

\begin{table}[H]
\centering
\caption{Persistent-minus-reset best-reward gain after the specified number of valid candidates. Values are normalized reward points; parentheses give paired task coverage.}
\label{tab:reward-by-candidate}
\begin{scriptsize}
\setlength{\tabcolsep}{5.0pt}
\renewcommand{\arraystretch}{1.08}
\begin{tabular}{@{}lrrr@{}}
\toprule
Agent & Candidate 1 & Candidate 2 & Candidate 3 \\
\midrule
Opus 5      & +0.7 (32) & -0.5 (32) & -0.4 (28) \\
GPT-5.6-Sol & -0.8 (32) & +0.6 (31) & -0.3 (29) \\
Kimi K3     & +1.7 (32) & +1.6 (29) & +0.5 (10) \\
Sonnet 4.6  & +6.8 (32) & +4.2 (28) & +12.2 (18) \\
GPT-5.4     & +3.7 (32) & +0.9 (32) & -1.0 (24) \\
\bottomrule
\end{tabular}
\end{scriptsize}
\end{table}

\paragraph{Improvement after the first valid solution.}
We also look at improvement after the first solution by substracting the first-valid reward from each trajectory. The improvement is non-positive for all agents (Table~\ref{tab:post-first-productivity}). However, Sonnet and Kimi produced a larger fraction of improving follow-up candidates, but their total first-to-best improvements are nearly identical between conditions. The current results therefore do not show that persistence generally improves optimization after initialization.

\begin{table}[H]
\centering
\caption{Productivity after the first valid solution over transfer-eligible tasks. Improving follow-ups are the fraction of later valid candidates that establish a new best. First-to-best improvement and incremental area gain are normalized reward points. S/R denotes stateful/reset.}
\label{tab:post-first-productivity}
\begin{scriptsize}
\setlength{\tabcolsep}{3.0pt}
\renewcommand{\arraystretch}{1.08}
\begin{tabular}{@{}lcrrr@{}}
\toprule
Agent & \makecell{Paired follow-up\\tasks} & \makecell{Improving\\follow-ups S/R} & \makecell{First-to-best\\improvement S/R} & \makecell{Incremental\\area gain} \\
\midrule
Opus 5      & 32 & 29.9\%/44.4\% & 1.8/3.3 & -1.0 \\
GPT-5.6-Sol & 31 & 31.5\%/36.8\% & 5.3/5.7 & -0.3 \\
Kimi K3     & 29 & 34.8\%/25.3\% & 0.5/0.6 & -0.2 \\
Sonnet 4.6  & 28 & 32.0\%/26.4\% & 4.2/4.1 & -0.2 \\
GPT-5.4     & 32 & 27.2\%/43.6\% & 1.1/5.3 & -3.2 \\
\bottomrule
\end{tabular}
\end{scriptsize}
\end{table}

\subsection{Effect of Stream Length} \label{app:stream-length}
We also run an ablation on how the number of prior tasks affects the whole stream. We see that streams with two, four, and six prior tasks improve initialization by \(+3.56\), \(+5.53\), and \(+5.34\) normalized reward points relative to a fresh run (Table~\ref{tab:task-count-ablation}). Final outcome gains remain near zero or negative, while time-convergence falls sharply with six prior tasks. 

\begin{table}[H]
    \centering
    \caption{\textbf{Effect of stream length on a common final Weather forecasting task.} Stream length includes the final task. GPT-5.4 gains are normalized reward points relative to a fresh run. First and selected scores are test RMSE, where lower is better. Each condition contains one run.}
    \label{tab:task-count-ablation}
    \small
    \setlength{\tabcolsep}{4pt}
    \renewcommand{\arraystretch}{1.12}
    \begin{tabular}{@{}crrrcc@{}}
        \toprule
        Stream length & \makecell{Init.\\gain} & \makecell{Time-conv.\\gain} & \makecell{Outcome\\gain} & \makecell{Valid\\attempts} & \makecell{First/selected\\RMSE} \\
        \midrule
        3 & \(+3.56\) & \(-1.75\) & \(-1.82\) & 3 & \(0.432 / 0.427\) \\
        5 & \(+5.53\) & \(-3.61\) & \(-0.05\) & 3 & \(0.393 / 0.392\) \\
        7 & \(+5.34\) & \(-31.44\) & \(-0.27\) & 2 & \(0.396 / 0.396\) \\
        \bottomrule
    \end{tabular}
\end{table}

\subsection{Operational Reliability} \label{app:operational-reliability}
We also notice that failure rates do not change uniformly under persistence.  Among valid and failed attempts on transfer-eligible tasks, persistence reduces the failure fraction from \(30.7\%\) to \(7.7\%\) for Sonnet and from \(11.4\%\) to \(1.8\%\) for Kimi. However, it revers for GPT-5.6-Sol, from \(1.7\%\) under reset execution to \(8.3\%\) under persistence. We also observe that the number of attempt reflect how much an agent decides to explore. The underlying valid and failed counts are reported in Table~\ref{tab:model-results-aggregate-appendix}.

\subsection{Transfer by Task Nature} \label{app:task-nature-analysis}
We also look at how transfer gains vary by task nature. We see that classification has the most consistent aggregate outcome transfer, with a \(60.0\%\) win rate and a positive median gain, while regression is evenly split. Symbolic regression has the largest mean outcome gain (\(+5.2\) points) but a negative median (\(-2.9\) points), showing that its average is driven by model-specific gains rather than broad improvement. Table~\ref{tab:task-nature-aggregate} reports the full comparison.

\subsection{Memory Contrasts} \label{app:memory-contrasts}

Table~\ref{tab:memory-ablation-model-results} reports the model-level memory contrasts.

\begin{table}[H]
\centering
\caption{Memory-ablation results on the four transfer-eligible tasks in the Energy-system stream, reported as mean [median] normalized reward points. M/S compares matched with shuffled cards; M/N compares matched cards with an independent no-memory reference. Positive values favor matched cards. Task 1 is excluded.}
\label{tab:memory-ablation-model-results}
\begin{scriptsize}
\setlength{\tabcolsep}{2.2pt}
\renewcommand{\arraystretch}{1.08}
\begin{tabular}{@{}
>{\raggedright\arraybackslash}p{0.185\linewidth}
>{\centering\arraybackslash}p{0.055\linewidth}
>{\centering\arraybackslash}p{0.155\linewidth}
>{\centering\arraybackslash}p{0.155\linewidth}
>{\centering\arraybackslash}p{0.155\linewidth}
>{\centering\arraybackslash}p{0.155\linewidth}
@{}}
\toprule
Model & Tasks & \makecell{Outcome\\gain M/S}
& \makecell{Outcome\\gain M/N}
& \makecell{Init.\\gain M/S}
& \makecell{Time-conv.\\gain M/S} \\
\midrule
Opus 5       & 4 & +1.0 [+0.7] & +0.2 [+0.4] & +2.6 [+2.7] & -15.5 [-13.2] \\
GPT-5.6-Sol  & 4 & -3.2 [-0.2] & +0.3 [-0.1] & +3.9 [+0.2] & -27.3 [-28.7] \\
Sonnet 4.6   & 4 & -0.2 [-1.2] & +14.7 [+8.3] & -2.9 [-1.5] & +25.6 [+26.8] \\
GPT-5.4      & 4 & +0.2 [-0.1] & +5.0 [+2.3] & +5.9 [+7.8] & -0.9 [-1.6] \\
\bottomrule
\end{tabular}
\end{scriptsize}
\end{table}

\paragraph{Sensitivity to the evaluation horizon.}

Table~\ref{tab:memory-budget-checkpoints} compares matched and shuffled memory at four budget checkpoints. The trend changes across checkpoints for Opus 5, Sonnet 4.6, and GPT-5.4. Relevant cards therefore change when progress occurs, but do not provide a consistent advantage at either shorter or longer horizons.

\begin{table}[H]
\centering
\caption{Matched-minus-shuffled difference in the best hidden reward achieved by each budget checkpoint on the four transfer-eligible Energy-system tasks. Values are normalized reward points; positive values favor matched memory. The best-so-far reward is zero before the first valid submission.}
\label{tab:memory-budget-checkpoints}
\begin{scriptsize}
\setlength{\tabcolsep}{7.0pt}
\renewcommand{\arraystretch}{1.08}
\begin{tabular}{@{}lrrrr@{}}
\toprule
Agent & 25\% & 50\% & 75\% & 100\% \\
\midrule
Opus 5      & -51.8 & -18.6 &  +3.0 & +1.6 \\
GPT-5.6-Sol & -14.1 & -53.3 & -35.3 & -3.6 \\
Sonnet 4.6  & +29.2 & +46.0 & +15.2 & -0.2 \\
GPT-5.4     & -15.3 &  +4.4 &  +0.3 & -0.1 \\
\midrule
Mean        & -13.0 &  -5.4 &  -4.2 & -0.6 \\
\bottomrule
\end{tabular}
\end{scriptsize}
\end{table}

Matched cards improve final outcome over no memory by \(+14.7\) points for Sonnet and \(+5.0\) for GPT-5.4, but shuffled cards show gains of \(+14.9\) and \(+4.8\) points. For Opus and GPT-5.6-Sol, matched-versus-no-memory outcome gains are near zero. The no-memory comparison therefore does not establish a semantic-relevance benefit beyond the matched-versus-shuffled contrast.

\section{Agent Trace Case Studies} \label{app:agent-traces}

We present three transfer-eligible cases in which prior context visibly changes a later search. Each figure contrasts the relevant conditions using parsed agent messages and observable actions; hidden reasoning is excluded. Hidden-test scores were computed after the run and were not visible to the agent. The cases are selected for interpretability and should not be read as a systematic or causal decomposition of the aggregate results.

\subsection{Direct Code and Search-Policy Reuse} \label{app:trace-code-reuse}

On the third ETT contract, stateful Sonnet recognized the same forecasting structure as the preceding hourly task. It reused its complete training pipeline but recalibrated ensemble weights and rejected an overtrained iTransformer when visible development performance worsened. The stateless agent instead built a seasonal model from a clean start (Figure~\ref{fig:trace-code-reuse}).

\begin{figure}[H]
\centering
\begingroup
\setlength{\fboxsep}{7pt}
\begin{minipage}[t]{0.485\textwidth}
\centering
\textbf{Stateful Sonnet reuses and adapts the pipeline}\par\smallskip
\fcolorbox{blue!55!black}{blue!4}{%
\parbox[t]{0.91\linewidth}{%
\raggedright\ttfamily\footnotesize
[Task transition]\par
``Same structure as the last task. Let me compare the actual data values to the previous task.''\par\medskip
``Different data---same structure, different values. I can reuse the exact same code.''\par\medskip
[Local checks]\par
Reuse DLinear, NLinear, and iTransformer pipeline. Refit all components, recalibrate ensemble weights, and drop the overtrained iTransformer.\par\medskip
[Hidden evaluation]\par
First = final RMSE: 0.5319
}}
\end{minipage}\hfill
\begin{minipage}[t]{0.485\textwidth}
\centering
\textbf{Stateless Sonnet rebuilds from a clean start}\par\smallskip
\fcolorbox{red!45!black}{red!3}{%
\parbox[t]{0.91\linewidth}{%
\raggedright\ttfamily\footnotesize
[Fresh session]\par
No earlier task transcript, code, fitted components, or development results.\par\medskip
[Independent search]\par
Construct an exponentially weighted same-hour seasonal model and tune its decay on the current task.\par\medskip
[Hidden evaluation]\par
First RMSE: 0.6249\par
Final RMSE: 0.5962\par\medskip
The fresh search improves, but does not catch the stateful first solution.
}}
\end{minipage}
\endgroup
\caption{\textbf{Direct reuse on a related ETT task.} The stateful agent explicitly reuses its earlier implementation, then recalibrates it using current-task evidence. Its first valid solution already outperforms the stateless agent's final solution.}
\label{fig:trace-code-reuse}
\end{figure}

This case is direct evidence of code and search-policy reuse. It does not show that Sonnet recovered a deeper relationship between the underlying series; the transferable object was the working pipeline and the procedure for adapting it.

\subsection{Matched Insights Change Initialization} \label{app:trace-matched-init}

For the Solar 10 Minutes task, fresh Codex sessions received either four cards extracted from preceding tasks in the same stream or four cards from unrelated forecasting streams. The matched cards supplied a useful abstraction, but Codex still had to correct the task-specific period (Figure~\ref{fig:trace-matched-init}).

\begin{figure}[H]
\centering
\begingroup
\setlength{\fboxsep}{7pt}
\begin{minipage}[t]{0.485\textwidth}
\centering
\textbf{Matched cards supply the right abstraction}\par\smallskip
\fcolorbox{blue!55!black}{blue!4}{%
\parbox[t]{0.91\linewidth}{%
\raggedright\ttfamily\footnotesize
[Read 4 matched cards]\par
``The prior notes suggest this task family may already have a strong linear residual solution, but I'm not assuming it transfers without checking the actual series structure.''\par\medskip
[Inspect current cadence]\par
``Daily structure is not at lag 96 here. The strongest simple periodicity is around lag 288.''\par\medskip
[Action]\par
Keep the seasonal-plus-residual idea; replace the inherited lag with a cadence-appropriate period and fit a low-rank residual ridge.\par\medskip
[Hidden RMSE]\par
First: 0.5449 \quad Final: 0.4125
}}
\end{minipage}\hfill
\begin{minipage}[t]{0.485\textwidth}
\centering
\textbf{Shuffled cards lead to a weaker starting point}\par\smallskip
\fcolorbox{red!45!black}{red!3}{%
\parbox[t]{0.91\linewidth}{%
\raggedright\ttfamily\footnotesize
[Read 4 size-matched shuffled cards]\par
Cards describe unrelated weekly-health and daily-currency forecasting tasks.\par\medskip
[Search]\par
Test generic linear and seasonal models, repair the 96-step window geometry, and later discover a rotated latent residual ridge.\par\medskip
[Hidden RMSE]\par
First: 1.4058 \quad Final: 0.4773\par\medskip
The broader search catches up substantially, but starts far behind matched memory.
}}
\end{minipage}
\endgroup
\caption{\textbf{Matched versus shuffled insight cards on Solar 10 Minutes.} Matched memory transfers an abstract seasonal-residual hypothesis while leaving its period open to verification. The largest difference appears in first-solution quality; full-budget search narrows the gap.}
\label{fig:trace-matched-init}
\end{figure}

This comparison shows why memory content matters at initialization even when final performance is less separated. Codex used the matched cards as a hypothesis to test, not as a fixed recipe.

\subsection{Specific Memory Can Create Anchoring} \label{app:trace-anchoring}

On Energy Transformer Beta, matched Sonnet received 21 cumulative cards, including notes closely matching the target schema. Shuffled Sonnet received the same number of cards from other streams. The matched condition produced a valid model earlier, but the specific prior narrowed its search (Figure~\ref{fig:trace-anchoring}).

\begin{figure}[H]
\centering
\begingroup
\setlength{\fboxsep}{7pt}
\begin{minipage}[t]{0.485\textwidth}
\centering
\textbf{Matched cards encourage a narrow correction model}\par\smallskip
\fcolorbox{red!45!black}{red!3}{%
\parbox[t]{0.91\linewidth}{%
\raggedright\ttfamily\footnotesize
[Read 21 matched cards]\par
``The prior insights describe this exact dataset (7-channel hourly energy, 8640 train rows). The best known result is a correction model at RMSE=0.837.''\par\medskip
[Action]\par
Reuse the seasonal-correction framing and add cross-load and temperature aggregates. Produce a valid model after 9.2 minutes.\par\medskip
[Hidden RMSE]\par
First = final: 0.5882
}}
\end{minipage}\hfill
\begin{minipage}[t]{0.485\textwidth}
\centering
\textbf{Shuffled cards permit broader exploration}\par\smallskip
\fcolorbox{blue!55!black}{blue!4}{%
\parbox[t]{0.91\linewidth}{%
\raggedright\ttfamily\footnotesize
[Read 21 size-matched shuffled cards]\par
No card closely identifies the target schema or a single preferred correction model.\par\medskip
[Action]\par
Explore more broadly and select a context-mean NLinear Ridge model with daily harmonics. First valid model arrives after 21.4 minutes.\par\medskip
[Hidden RMSE]\par
First = final: 0.5431\par\medskip
No-memory stateless reference: 0.5379
}}
\end{minipage}
\endgroup
\caption{\textbf{Earlier availability does not guarantee a better final model.} Highly specific matched cards make Sonnet productive sooner but anchor it to a correction-model family. Shuffled cards delay the first valid solution yet lead to broader search and a better final result.}
\label{fig:trace-anchoring}
\end{figure}

This case illustrates a failure mode of explicit memory: close surface similarity can increase confidence in a prior without ensuring that its empirical advantage survives a new split. Memory mechanisms therefore need support for verification and rejection, not only retrieval.

\section{Agent Failure Case Studies} \label{app:agent-failures}
To understand why, in certain cases, agents are not able to improve the final results while showing signs of better search. We present different cases that highlights different failures in search policy, model selection, and operational adaptation. 

\subsection{Dev Optimization Can Erase a Transfer Advantage} \label{app:failure-symbolic-selection}
In one of Kimi K3 stateful canonical-coordinates task, it carried forward a search procedure based on previous feature screening, handcrafted basis construction, and greedy refinement. This policy seemed to have helped early on by producing a useful expression early, but subsequent optimization moved the selected expression away from the candidate that ended generalizing the best under hidden extrapolation (Figure~\ref{fig:failure-symbolic-selection}).

\begin{figure}[H]
\centering
\begingroup
\setlength{\fboxsep}{7pt}
\begin{minipage}[t]{0.485\textwidth}
\centering
\textbf{An earlier expression extrapolates well}\par\smallskip
\fcolorbox{blue!55!black}{blue!4}{%
\parbox[t]{0.91\linewidth}{%
\raggedright\ttfamily\footnotesize
[Structure search]\par
``Breakthrough: $\sin(x_{011}x_{012})$ alone gives dev 0.960!''\par\medskip
[Second valid expression]\par
Retain the sine-product structure with a small set of corrections.\par\medskip
Development $R^2$: 0.961\par
Hidden $R^2$: 0.593\par
Found after 7.3 minutes.
}}
\end{minipage}\hfill
\begin{minipage}[t]{0.485\textwidth}
\centering
\textbf{Dev polishing selects a fragile expression}\par\smallskip
\fcolorbox{red!45!black}{red!3}{%
\parbox[t]{0.91\linewidth}{%
\raggedright\ttfamily\footnotesize
[Residual search]\par
Add quadratic, trigonometric, and absolute-value corrections over several additional variables.\par\medskip
``Dev R2 0.99979! The full additive structure found. Promoting to final immediately.''\par\medskip
[Selected expression]\par
Development $R^2$: 0.99979\par
Hidden $R^2$: $-1.354$\par
Selection regret: 41.3 reward points.
}}
\end{minipage}
\endgroup
\caption{\textbf{Within-task optimization destroys a strong transferred initialization.} Stateful Kimi finds its best hidden-generalizing expression early, but selects a more elaborate expression because it nearly saturates visible development $R^2$. The additional terms fail under hidden extrapolation.}
\label{fig:failure-symbolic-selection}
\end{figure}

However, this failure was not due to Kimi not learning across tasks. In fact, Kimi maintained a work log, reused its symbolic search tools, and reached a very good solution quickly. However, what caused the failure was the inductive bias it learned through that procedure. Small development improvements favor flexible approximations even when the task requires recovery of a compact generating structure. Also, the agent overwrote a good candidate instead of using extrapolation robustness as a separate selection objective. This example illustrate that better agents should retain a portfolio of structurally distinct candidates, penalize ungrounded complexity, and stress-test expressions outside what is observed. 

\subsection{Finding a Strong Candidate Is Not Enough} \label{app:failure-forecast-selection}
In one of GPT-5.6-Sol Energy Transformer Beta run, it explicitly reused the previous task's forecasting solution while refitting all parameters to it. It was able to find a competitive model within five minutes and continued to improve its model. However, the final selection generalized much worse than both earlier attempts (Figure~\ref{fig:failure-forecast-selection}).

\begin{figure}[H]
\centering
\begingroup
\setlength{\fboxsep}{7pt}
\begin{minipage}[t]{0.485\textwidth}
\centering
\textbf{Transfer accelerates candidate generation}\par\smallskip
\fcolorbox{blue!55!black}{blue!4}{%
\parbox[t]{0.91\linewidth}{%
\raggedright\ttfamily\footnotesize
[Task transition]\par
``Same schema but a distinct Beta series, so I'll reuse only the proven modeling machinery---not its fitted state or weights.''\par\medskip
[Task-local adaptation]\par
Refit every component, retune analog neighbors, and remove a tree component that weakens the chronological holdout.\par\medskip
[Second valid candidate]\par
Dev RMSE: 0.453\par
Hidden RMSE: 0.545\par
Found after 4.5 minutes.
}}
\end{minipage}\hfill
\begin{minipage}[t]{0.485\textwidth}
\centering
\textbf{Dev selection promotes the wrong final model}\par\smallskip
\fcolorbox{red!45!black}{red!3}{%
\parbox[t]{0.91\linewidth}{%
\raggedright\ttfamily\footnotesize
[Continued refinement]\par
Expand the ridge grid and repeatedly recalibrate stacking weights. Each accepted change improves visible dev RMSE.\par\medskip
``The final adjacent grid also improves: official RMSE is now 0.417466. I'm promoting attempt 0010.''\par\medskip
[Hidden evaluation]\par
Selected stateful RMSE: 1.049\par
Earlier stateful candidate: 0.545\par
Reset final: 0.523
}}
\end{minipage}
\endgroup
\caption{\textbf{Search succeeds, but final selection fails.} Persistent state helps GPT-5.6-Sol find a useful candidate much earlier, yielding a $+55.0$-point time-convergence gain. Continued dev optimization selects a fragile ensemble, however, producing a $-16.9$-point final outcome gain.}
\label{fig:failure-forecast-selection}
\end{figure}

We note that the agent did not blindly reuse the model from its previous task. It refitted all components and validated a chronological holdout. Yet, the holdout was insufficient to predict the hidden regime. The main limitation for the agents is then selection robustness rather than generating good candidates. However, a continual agent should preserve its best simple incumbent, evaluate replacements across multiple time splits and conditions. Thus, candidate generation and final selection should be treated as separate research decisions.

\subsection{A Correct Operational Lesson Can Be Applied Too Strongly} \label{app:failure-defensive-policy}
In one of Opus 5 MLE-Bench run, it launched a large cross-validation run before creating a valid artifact. At the end once the time budget was over, there was only an empty final directory. The agent, however, correctly diagnoses the problem and writes a persistent safe-first policy. That policy prevents another null delivery, but its conservative application reduces exploration and final quality on both subsequent tasks (Figure~\ref{fig:failure-defensive-policy}).

\begin{figure}[H]
\centering
\begingroup
\setlength{\fboxsep}{7pt}
\begin{minipage}[t]{0.485\textwidth}
\centering
\textbf{The agent learns from a delivery failure}\par\smallskip
\fcolorbox{blue!55!black}{blue!4}{%
\parbox[t]{0.91\linewidth}{%
\raggedright\ttfamily\footnotesize
[Task 1]\par
Launch a five-model, five-fold CV suite before writing a final artifact. The process reaches the deadline; no valid submission is produced.\par\medskip
[Postmortem]\par
``I launched a 5-model CV suite as my first action and blocked waiting on it. The deadline passed with final empty.''\par\medskip
[Persistent policy]\par
``Write a valid final within the first $\sim$3 minutes, before any tuning.'' Also cap threads and bound later experiments.
}}
\end{minipage}\hfill
\begin{minipage}[t]{0.485\textwidth}
\centering
\textbf{The safe-first policy becomes too conservative}\par\smallskip
\fcolorbox{red!45!black}{red!3}{%
\parbox[t]{0.91\linewidth}{%
\raggedright\ttfamily\footnotesize
[Task 2]\par
``Lesson applied: ship a valid final immediately, then improve.'' Train a bounded LightGBM on a 150k-row subset and produce only two valid attempts.\par\medskip
[Reset comparison]\par
Engineer the same character signal, train and rank-blend 11 MLPs across seeds and splits, and produce eight valid submissions.\par\medskip
[Hidden ROC-AUC]\par
Stateful: 0.978\par
Reset: 0.998\par
Time-convergence gain: $-85.4$ points.
}}
\end{minipage}
\endgroup
\caption{\textbf{Operational learning improves validity but sacrifices exploration.} Opus converts a real failure into a reusable policy and avoids another missing submission. Applied without an adaptive exploration budget, however, the policy protects a weak fallback while the reset agent searches more broadly and finishes substantially better.}
\label{fig:failure-defensive-policy}
\end{figure}

Host contention also affected this stream, so the timing gap cannot be attributed to memory alone. The policy change, however, is observable: later stateful tasks prioritized bounded subsamples and protected fallbacks, yielding 4 valid attempts across the stream compared with 28 in reset runs. A better policy would separate reliability from exploration by first securing an initial artifact, then allocating the remaining budget to alternatives with higher potential gains. Operational memories should record the conditions under which they were formed and the criteria for revising them, rather than functioning as unconditional rules.

\end{document}